\documentclass[10pt,twocolumn,letterpaper]{article}

\usepackage{cvpr}              

\newcommand{\TODO}[1]{\textbf{\color{red}[TODO: #1]}}
\renewcommand{\TODO}[1]{}

\definecolor{cvprblue}{rgb}{0.21,0.49,0.74}
\usepackage[pagebackref,breaklinks,colorlinks,allcolors=cvprblue]{hyperref}

\def\paperID{17627} 
\def\confName{CVPR}
\def\confYear{2026}

\title{FlashLips: 100-FPS Mask-Free Latent Lip-Sync using Reconstruction Instead of Diffusion or GANs}

\author{
Andreas Zinonos\textsuperscript{1*} \quad
Michał Stypułkowski\textsuperscript{2} \quad
Antoni Bigata\textsuperscript{1,3} \\
Stavros Petridis\textsuperscript{1,3} \quad
Maja Pantic\textsuperscript{1,3} \quad
Nikita Drobyshev\textsuperscript{2*} \\ 
\textsuperscript{1}Imperial College London \quad
\textsuperscript{2}Cantina Labs  \quad
\textsuperscript{3}NatWest AI Research \\
{\tt\small \{andreas.zinonos18, a.bigata-casademunt22, stavros.petridis04, m.pantic\}@imperial.ac.uk} \\
{\tt\small \{nikita, michal.stypulkowski\}@cantina.ai}
}

\begin{document}
\twocolumn[{%
\renewcommand\twocolumn[1][]{#1}%
\maketitle
\begin{center}
    \centering
    \captionsetup{type=figure}
    \includegraphics[width=0.9\textwidth]{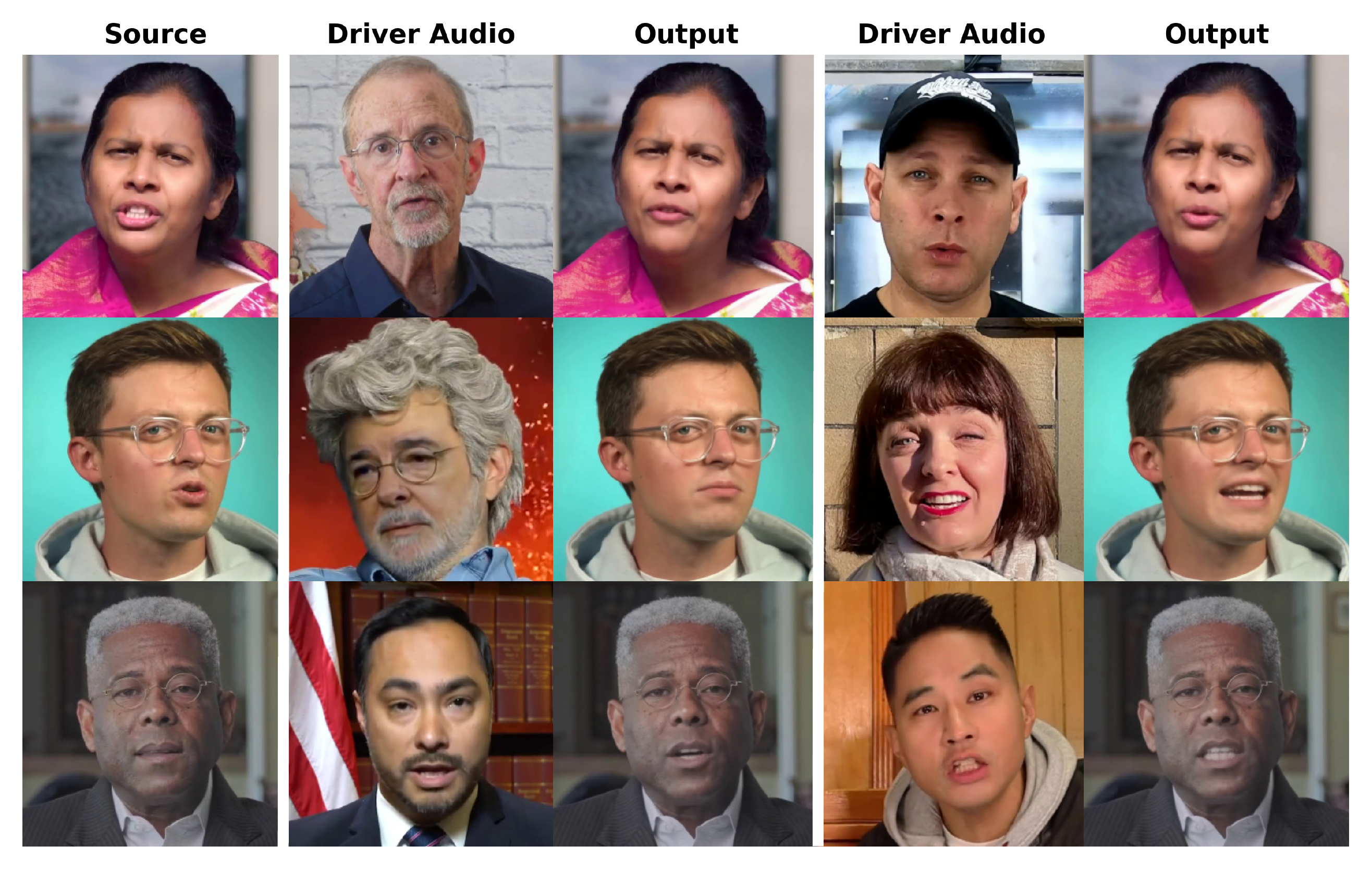}
    \caption{\textbf{FlashLips Results.} Selected results of source and driver pairs, generated using our transformer-based model.}
\end{center}%
}]

\begingroup
\renewcommand\thefootnote{$*$}
\footnotetext{Equal Contribution}
\endgroup


\begin{abstract}
We present \emph{FlashLips}, a two-stage, mask-free lip-sync system that decouples lips control from rendering and achieves real-time performance, with our U-Net variant running at over 100 FPS on a single GPU, while matching the visual quality of larger state-of-the-art models. Stage~1 is a compact, one-step latent-space editor that reconstructs an image using a reference identity, a masked target frame, and a low-dimensional lips-pose vector, trained purely with reconstruction losses -- no GANs or diffusion. To remove explicit masks at inference, we use self-supervision via mouth-altered target variants as pseudo ground truth, teaching the network to localize lip edits while preserving the rest. Stage~2 is an audio-to-pose transformer trained with a flow-matching objective to predict lips-pose vectors from speech. Together, these stages form a simple and stable pipeline that combines deterministic reconstruction with robust audio control, delivering high perceptual quality and faster-than-real-time speed.
\end{abstract}

\section{Introduction}
\label{sec:intro}
Lip synchronization (lip-sync) is the task of regenerating realistic mouth movements that match audio while preserving identity, expression, head pose, background, and overall fidelity of a talking-person video. It has a transformative impact across domains — from automating film/TV dubbing and breaking language barriers, to creating expressive animations and lifelike digital avatars \cite{zhang2023dinet, zhen2023}. The central challenge is synthesizing photorealistic, temporally stable lip motions precisely synchronized with speech.

Audio-driven facial generation is closely related: it animates the full face from a reference image and a driving audio \cite{zhou2019talking, vougioukas2019gans, xu2024hallo, chen2024echomimic, wang2024vexpress}. Compared to full-face generation, lip-sync is more controllable and efficient: it edits only the mouth while reusing identity, pose, and background from the target video, crucial for dubbing/localization.

The pursuit of high fidelity has produced a spectrum of deep learning approaches \cite{goodfellow2016deep}. Early successes were driven by GANs \cite{goodfellow2014gans, prajwal2020wav2lip, guan2023stylesync}, which can yield sharp frames but are notoriously difficult to train and sensitive to hyperparameters \cite{Salimans2016, Srivastava2017}. More recently, iterative generative models -- particularly diffusion -- have set a strong bar for visual quality in both general face animation \cite{shen2023difftalk, stypulkowski2024diffused, zhang2024dreamtalk, du2025rap} and task‑specific lip‑sync \cite{bigioi2024, liu2024diffdub, mukhopadhyay2024}. However, diffusion requires sequential inference (multiple denoising steps), compounding cost and often prompting additional pre/post-processing such as explicit mouth masks or alignment to canonical templates \cite{li2025latentsync, bigata2025keysync}, which complicates real-time deployment and adds engineering overhead.

In this work, we take a step back and question the necessity of iterative \emph{visual} generators for a highly conditioned task like lip‑sync. We argue that with sufficient context -- a reference identity, a target frame, and precise lips‑pose cues -- a powerful \emph{deterministic} image update can be learned without adversarial objectives or diffusion schedules. 

We introduce a lightweight, two‑stage framework that separates \emph{control} from \emph{rendering} in the spirit of two‑stage designs of \cite{liu2024diffdub}.

\begin{figure}[t]
  \centering
  \includegraphics[width=\columnwidth]{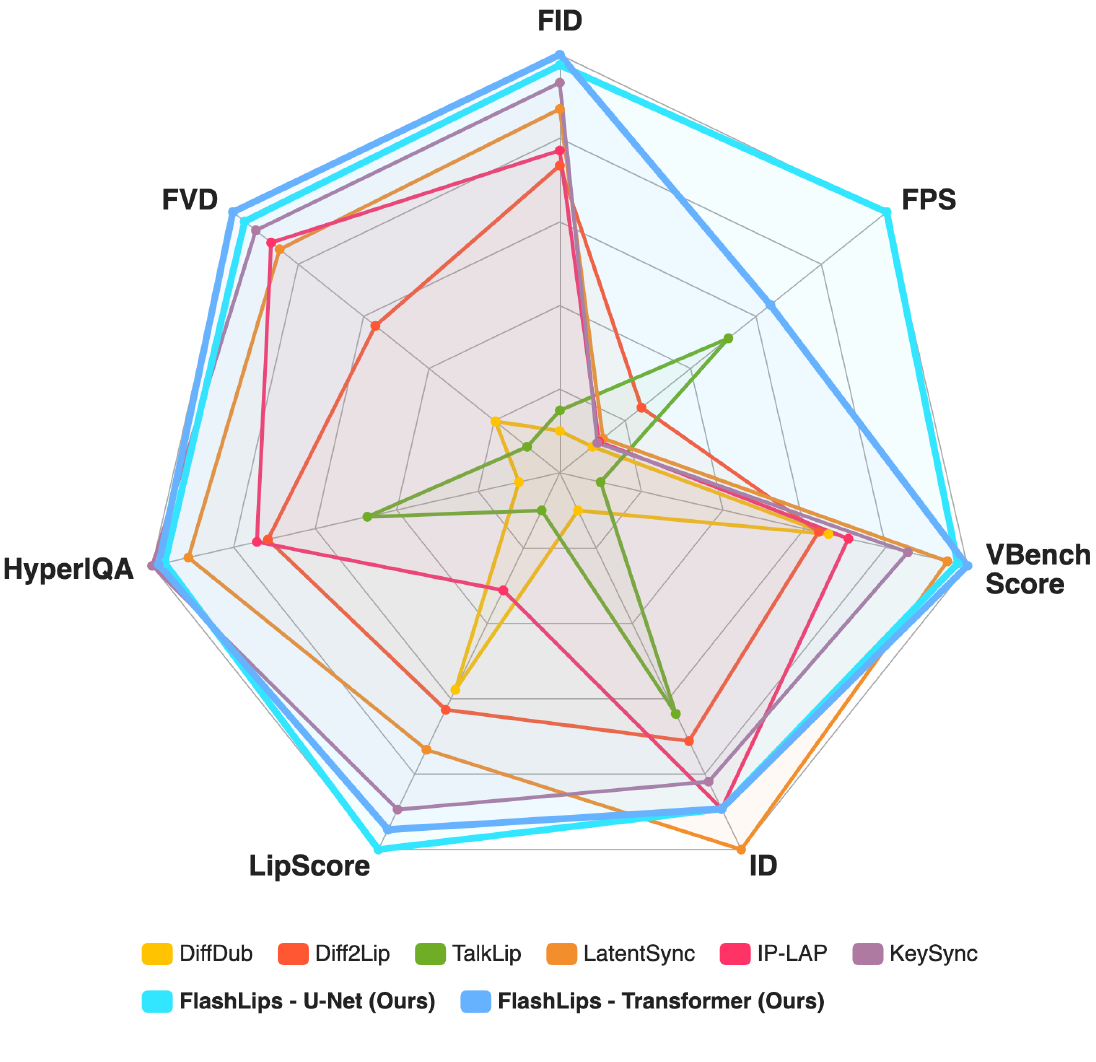}
  \vspace{-1em}
    \caption{\textbf{Visualization of Quantitative Evaluation.} Comparison of eight lip-sync models in the cross-audio setting on seven key metrics. Results are normalized, with the best-performing model scaled to the outer edge, and the worst towards the center.}
  \label{fig:cross_audio_radar}
  \vspace{-2em}
\end{figure}

\textbf{Stage 1: Latent Visual Editor.}
We start with a compact \emph{one‑step} editor operating in VAE latent space \cite{podell2023sdxl}. Given a reference image, a target frame, and a low‑dimensional \emph{lips‑pose vector}, it reconstructs the edited frame in a single feed‑forward pass using \emph{reconstruction losses only} - no adversarial training or diffusion.

Our final editor runs \emph{without} explicit mouth masks at inference. After reconstruction training, we self-refine by synthesizing mouth-altered variants and fine-tuning on symmetric \emph{source $\leftrightarrow$ changed} pairs, which focuses edits on the lips while preserving the rest of the frame without any external segmentation. Prior work established mask-free feasibility \cite{peng2025omnisync}; we instead create the supervision \emph{on-the-fly} with the editor itself, avoiding dataset-constructed pseudo-pairs.

\textbf{Stage 2: Audio‑to‑Lips.}
\emph{Stage~2} connects audio to the visual editor via an audio‑to‑lips transformer that predicts lips‑pose vectors from speech. A key design principle is to \emph{disentangle} the control space so that it carries \emph{pose} information only, \ie what the lips should do, while appearance (teeth, lip color, skin tone, jawline) and scene details are sourced from the reference and target frames in Stage~1. This mirrors what can be reliably inferred from audio and keeps Stage~2 lightweight and stable to train. Overloading audio with appearance factors makes learning harder and harms generalization; our disentangled control avoids this by construction. Conditioning on wav2vec 2.0 features \cite{baevski2020wav2vec} and training with \emph{flow-matching} \cite{lipman2022flow, liu2022flow} yields smooth control latents that drive the editor.

\vspace{0.5em}
\noindent\textbf{Our contributions}:
\begin{itemize}
  \item \textbf{Real-time performance:} $>$100 FPS on a single NVIDIA H100 80GB HBM3 with our U-Net variant, that matches or exceeds larger, slower baselines in terms of lip-sync accuracy and quality (\cref{fig:cross_audio_radar}).
  \item \textbf{Deterministic, one-step feasibility:} For a highly conditioned task such as lip-sync, where the target output is largely determined by the inputs, reconstruction-only training might be sufficient, removing the need for adversarial or iterative generators -- \emph{no GANs or diffusion}.
  \item \textbf{Mask-free self-refinement:} no explicit mouth masks at inference; fewer mouth artifacts and a simpler pipeline.
  \item \textbf{Disentangled audio-to-pose:} flow-matching transformer on wav2vec 2.0 separates \emph{what} to render from \emph{how} to render, supporting modular, per-component control.
\end{itemize}

\section{Related Work}
\label{sec:related}
\subsection{Audio-driven Portrait Animation}
\label{sec:rel_portrait_animation}
Audio-driven portrait animation synthesizes talking-head videos from speech~\cite{tian2025, Ji_2025_CVPR, jiang2025loopy}, yet its objectives differ fundamentally from lip synchronization. It follows an image-to-video paradigm that freely modifies head pose and facial expressions without constraining the output to match a specific input video. In contrast, lip synchronization operates in a video-to-video setting, modifying only the mouth region while preserving all other facial details, effectively editing the original video.

Early GAN-based approaches employed temporal and task-specific discriminators to animate faces~\cite{vougioukas2018e2e, vougioukas2019gans, zhou2019talking}, later incorporating head-pose modeling to improve the results, though often introducing visual artifacts~\cite{Zhou_2021_CVPR, Zhang_2023_CVPR}. Diffusion-based methods further enhanced temporal coherence and perceptual quality~\cite{stypulkowski2024diffused, jiang2025loopy, xu2024hallo}, sometimes leveraging facial landmarks or 3D meshes~\cite{wei2024aniportrait, zhang2023dreamtalk}, although the latter can lead to unrealistic animations. Frameworks with a two stage inference~\cite{bigata2025keyface} first generate intermediate keyframes and then interpolate between them to achieve smoother, temporally consistent motion. Despite producing plausible facial motion, many of these models suffer from high inference latency. To address this, recent works~\cite{xu2024vasa, ki2025float} train small audio-to-latent diffusion models that drive pre-trained latent-to-video decoders, achieving near real-time performance.

Although these methods have advanced portrait animation, their freedom to alter head pose and expression makes them unsuitable for lip synchronization, motivating dedicated methods that preserve identity and facial consistency.

\subsection{Audio-driven Lip Synchronization}
\label{sec:rel_lipsync}
Lip synchronization aims to modify only the mouth region to match the input audio while preserving head pose, identity, and other facial expressions. Wav2Lip~\cite{prajwal2020wav2lip} popularized SyncNet-based~\cite{chung2017syncnet} supervision to ensure reliable audio–visual alignment. Subsequent works such as DINet~\cite{zhang2023dinet}, IP-LAP~\cite{Zhong_2023_CVPR}, and ReSyncer~\cite{guan2025} enhance realism through spatial deformation of reference features, intermediate landmark prediction, or the use of 3D priors. Other works such as StyleSync~\cite{guan2023stylesync} and StyleLipSync~\cite{Ki_2023_ICCV} adopt StyleGAN-inspired architectures~\cite{Karras_2020_CVPR}, while TalkLip~\cite{wang2023seeing} leverages a lip-reading expert within a contrastive learning framework to enhance lip–speech synchronization.

Diffusion-based approaches further enhance temporal consistency and perceptual quality~\cite{mukhopadhyay2024, liu2024diffdub, bigioi2024, peng2025omnisync}. Latent diffusion frameworks such as LatentSync~\cite{li2025latentsync} and SayAnything~\cite{ma2025sayanything} synthesize lip-synced frames directly from audio without intermediate motion conditioning. MuseTalk~\cite{zhang2025musetalk} refines synchronization by selecting reference frames with similar head poses, and KeySync~\cite{bigata2025keysync} mitigates temporal inconsistency and leakage through a keyframe–interpolation approach and careful masking strategy.

Nevertheless, existing approaches still face notable limitations. GAN-based methods often suffer from visual artifacts and unstable training, while diffusion-based models, although capable of high-quality synthesis, remain computationally expensive and generally too slow for real-time inference. Moreover, many pipelines depend on extensive pre-processing (e.g., face alignment or intermediate motion estimation), which can introduce artifacts and reduce flexibility. These challenges motivate our work: a simple and efficient framework for high-quality lip synchronization that eliminates iterative generation and heavy pre-processing, enabling faster-than-real-time inference with high-resolution outputs.
\section{Method}
\label{sec:method}
We propose a two-stage framework for lip synchronization (\cref{fig:overview}). \textbf{Stage~1} is a fast, deterministic editor that produces high‑quality lip‑synced frames in a single forward pass. \textbf{Stage~2} is an audio‑to‑pose transformer that predicts low‑dimensional lip poses from speech and drives Stage~1. Stage~1 is trained \emph{only} with reconstruction objectives (no adversarial training or diffusion); Stage~2 uses a flow‑matching objective.

\subsection{Stage 1: Latent Visual Editor}
\subsubsection{Reconstruction and Lips Encoder}
\label{sssec:overview}
Stage~1 is trained per frame in latent space following \cite{rombach2022}. Let $\mathbf{x}_{\text{src}}, \mathbf{x}_{\text{ref}} \in \mathbb{R}^{C \times H \times W}$ be the frame to edit (``source'') and a reference frame from the same video, sampled $t$ frames apart, respectively. During reconstruction training we apply a mouth‑region mask to $\mathbf{x}_{\text{src}}$, yielding $\mathbf{x}_{\text{masked}}$.

We encode $\mathbf{x}_{\text{src}}$, $\mathbf{x}_{\text{masked}}$, and $\mathbf{x}_{\text{ref}}$ with the SDXL VAE \cite{kingma2014vae, podell2023sdxl} to obtain latents
$\mathbf{z}_{\text{src}}, \mathbf{z}_{\text{masked}}, \mathbf{z}_{\text{ref}} \in \mathbb{R}^{C_\ell \times H_\ell \times W_\ell}$.
A small trainable \emph{reference} backbone $f_{\text{ref}}$ projects the reference latent,
$\overline{\mathbf{z}}_{\text{ref}} = f_{\text{ref}}(\mathbf{z}_{\text{ref}})$,
so the model can better adapt identity features. We also use a lips‑pose representation $\mathbf{z}_{\text{lips}} \in \mathbb{R}^{M}$ that encodes lips configuration. It is expanded spatially by replication,
$\mathbf{z}_{\text{lips expanded}} \in \mathbb{R}^{M \times H_\ell \times W_\ell}$.

The network's input is the channel‑wise concatenation:
\begin{equation}
\mathbf{z}_{\text{input}}=\operatorname{Concat}\big[
\mathbf{z}_{\text{masked}},\;
\overline{\mathbf{z}}_{\text{ref}},\;
\mathbf{z}_{\text{lips expanded}}\big].
\end{equation}
It predicts a latent residual towards the ground‑truth edit. With:
\begin{equation}
\mathbf{z}_{\text{target}} = \mathbf{z}_{\text{src}} - \mathbf{z}_{\text{masked}},
\end{equation}
the model outputs $\hat{\mathbf{z}}_{\text{target}}$ and forms:
\begin{equation}
\hat{\mathbf{z}}_{\text{src}} = \mathbf{z}_{\text{masked}} + \hat{\mathbf{z}}_{\text{target}}.
\end{equation}
Decoding with the frozen VAE yields
$\hat{\mathbf{x}}_{\text{src}} = \text{VAE}_{\text{decoder}}(\hat{\mathbf{z}}_{\text{src}})$.

\begin{figure*}[t]
    \centering
    \includegraphics[width=\textwidth]{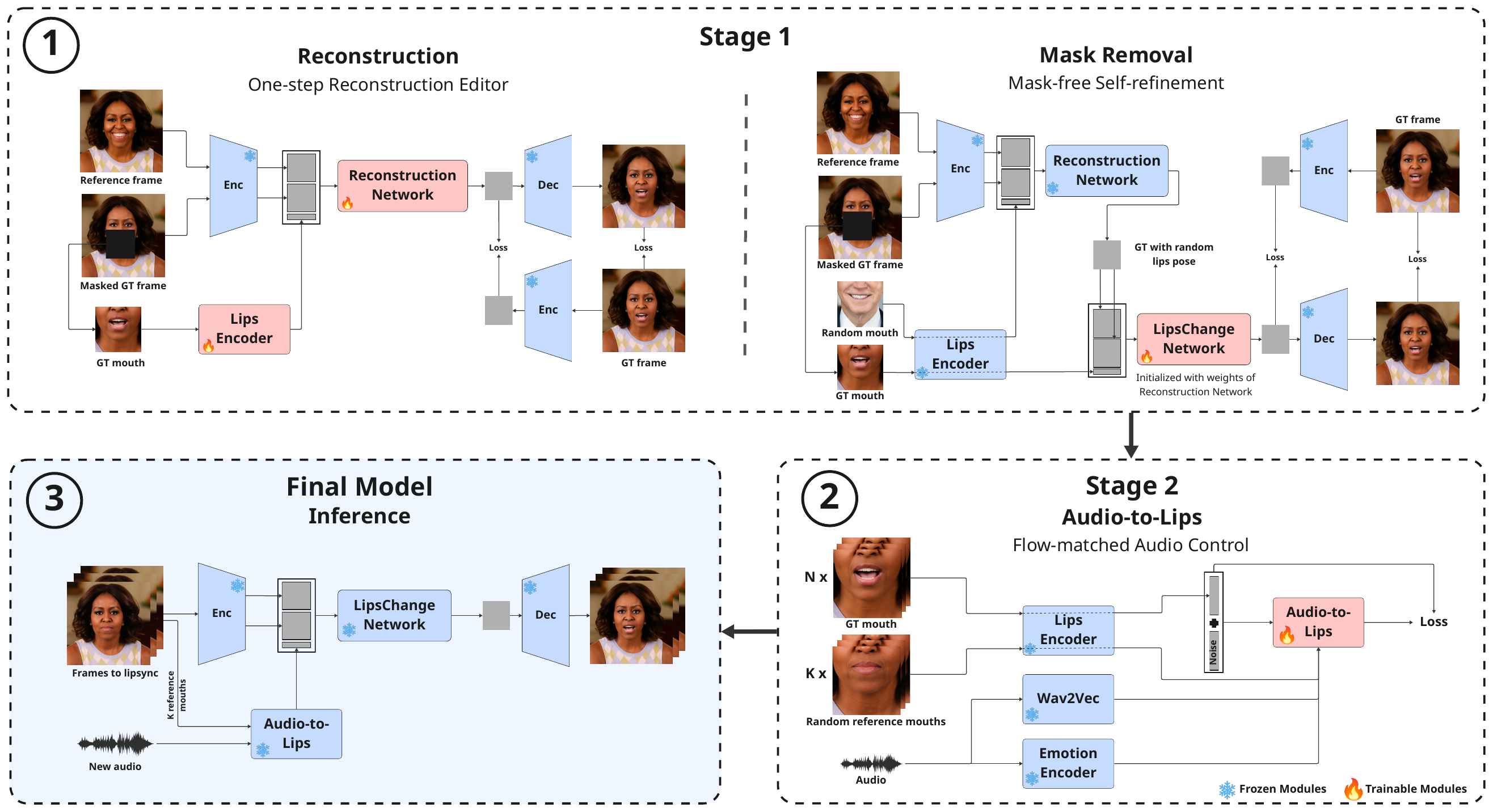}
    \caption{\textbf{Overview of FlashLips}. \textbf{Stage~1} trains a one-step latent-space editor: first via masked reconstruction, then via a mask-free self-refinement step that learns to localize edits without segmentation. \textbf{Stage~2} trains an audio-to-lips model that predicts the lips-pose vector used in Stage~1. At inference, predicted lip poses drive the LipsChange network to produce lip-synced frames in a single pass.
    }
    \label{fig:overview}
\end{figure*}

\noindent\textbf{Lips-Pose Representation.}
We design the representation to carry \emph{only} lip/jaw configuration to predict it more easily from audio. A frozen expression encoder \cite{drobyshev2024emoportraits} with a small MLP yields $\mathbf{z}_{\text{lips}}^{\text{main}}\!\in\!\mathbb{R}^{M}$. In parallel, a lightweight CNN on a mouth crop predicts a small residual $\mathbf{z}_{\text{lips}}^{\text{add}}$. The final control is
$\mathbf{z}_{\text{lips}}=\mathbf{z}_{\text{lips}}^{\text{main}}+\mathbf{z}_{\text{lips}}^{\text{add}}$.
For faster inference, we distill these two models into a compact image encoder (e.g., a ResNet-34 head \cite{He_2016_CVPR}) that predicts $\mathbf{z}_{\text{lips}}$ directly from an RGB face crop; see \cref{fig:lips_encoder} and the supplementary. Stage~2 is trained to predict the same vector from audio.

\subsubsection{Mask Removal via Self-Refinement}
\label{sssec:maskfree}
Once the reconstruction model converges, we sample lips vectors and synthesize lip-altered variants of the original frames to create symmetric pseudo-pairs
$(\text{source}\!\rightarrow\!\text{changed})$ and $(\text{changed}\!\rightarrow\!\text{source})$. A \emph{LipsChange} network initialized from the reconstruction model weights is then fine-tuned on these pseudo-pairs. This teaches the model to localize edits to the lips and preserve surrounding regions, eliminating the need for external segmentation (\cref{fig:overview}).

\subsubsection{Losses}
\label{sssec:losses}
Let $\mathbf{M}$ be the lower‑face pixel mask, and $\mathbf{m}$
its latent version downsampled with VAE. 
Lips mask $\mathbf{M}_{\text{lips}}$ comes from face parsing. 

 Let $\operatorname{MAE}_{\mathbf{w}}(A)=\frac{1}{|\Omega_{\mathbf{w}}|}\,\|\mathbf{w}\odot A\|_1$ and $\operatorname{MAE}(A)=\frac{1}{|\Omega|}\,\|A\|_1$, where $\Omega_{\mathbf{w}}$ indicates a support after applying the mask $\mathbf{w}$, and $\odot$ is the element-wise (Hadamard) product.

Let $\Delta\mathbf{z}=\hat{\mathbf{z}}_{\text{target}}-\mathbf{z}_{\text{target}}$ be the difference between predicted and ground-truth latents. Then, the losses in the latent space are defined as:
\begin{equation}
\label{eq:loss_lat}
\mathcal{L}^{lat}_{L1} = \operatorname{MAE}(\Delta\mathbf{z}),\quad
\mathcal{L}^{lat}_{L1_m} = \operatorname{MAE}_{\mathbf{m}}(\Delta\mathbf{z}).
\end{equation}

We also use the following losses in the pixel space:
\begin{align}
\label{eq:loss_pix}
\mathcal{L}^{pix}_{L1_M}      &= \operatorname{MAE}_{\mathbf{M}}\!\big(\hat{\mathbf{x}}_{\text{src}}-\mathbf{x}_{\text{src}}\big),\\
\mathcal{L}^{pix}_{L1_{\text{lips}}}
&= \mathds{1}_{\{|\Omega_{\text{lips}}|\ge \tau_{\text{lips}}\}}\,
   \operatorname{MAE}_{\mathbf{M}_{\text{lips}}}\!\big(\hat{\mathbf{x}}_{\text{src}}-\mathbf{x}_{\text{src}}\big),\\
\mathcal{L}_{VGG}             &= \sum_{l}\operatorname{MAE}\!\left(\phi_l(\hat{\mathbf{x}}_{\text{src}})-\phi_l(\mathbf{x}_{\text{src}})\right),\\
\mathcal{L}^{face}_{VGG}      &= \sum_{l}\operatorname{MAE}\!\left(\psi_l(\hat{\mathbf{x}}_{\text{src}})-\psi_l(\mathbf{x}_{\text{src}})\right),
\end{align}
where $\hat{\mathbf{x}}_{\text{src}}$ and $\mathbf{x}_{\text{src}}$ are predicted and ground-truth images, respectively, $\mathcal{L}_{VGG}$ uses VGG-19 features as in \cite{johnson2016vgg}, and $\mathcal{L}^{face}_{VGG}$ uses a VGGFace2-pretrained network \cite{cao2018vggface2datasetrecognisingfaces}.  
The lips loss is applied only when a valid lips mask is found and its area exceeds $\tau_{\text{lips}}$.

Finally, the total loss becomes:
\begin{equation}
\label{eq:total_loss}
\begin{split}
\mathcal{L}_{\text{total}} =\;&
0.1\,\mathcal{L}^{lat}_{L1}
+0.1\,\mathcal{L}^{lat}_{L1_m}
+10\,\mathcal{L}^{pix}_{L1_M} \\
&+100\,\mathcal{L}^{pix}_{L1_{\text{lips}}}
+50\,\mathcal{L}_{VGG}
+5\,\mathcal{L}^{face}_{VGG}.
\end{split}
\end{equation}

\subsection{Stage 2: Audio-to-Lips with Flow Matching}
\label{sec:stage2}
Stage~2 predicts the lips vector from speech and drives the editor trained in Stage~1. The model is a transformer conditioned on wav2vec 2.0 features \cite{baevski2020wav2vec}. We train it with a flow-matching objective \cite{lipman2022flow, liu2022flow} in the space of lips vectors.

Let $\boldsymbol{a}$ be the audio features aligned to a video frame, and $\mathbf{z}_{\text{mouth}}$ the target lips vector. We follow the optimal transport conditional flow-matching. We sample $\boldsymbol{\epsilon} \sim \mathcal{N}(0, \mathbf{I})$ and $t \sim \mathcal{U}(0,1)$, and define an interpolated point in latent space
\begin{equation}
    \mathbf{z}_t = (1-t)\,\boldsymbol{\epsilon} + t\,\mathbf{z}_{\text{lips}},
\end{equation}
and the target velocity
\begin{equation}
    \mathbf{u} = \mathbf{z}_{\text{lips}} - \boldsymbol{\epsilon}.
\end{equation}

The transformer $v_\theta$ is trained to match this velocity field,
\begin{equation}
    \mathcal{L}_{\text{FM}} = \mathbb{E}_{t,\,\boldsymbol{\epsilon},\,\boldsymbol{a}}\,
    \big\|\,v_\theta(\mathbf{z}_t, t, \mathbf{c}) - \mathbf{u}\,\big\|_2^2,
\end{equation}
where $\mathbf{c} = \text{Concat}[\boldsymbol{a}, e(\boldsymbol{a}), \mathbf{z}_{\text{lips}}^K]$, $e(\boldsymbol{a})$ is a pre-trained audio emotion encoder, and $\mathbf{z}_{\text{lips}}^K$ are $K$ randomly sampled source lip latents.

At inference, we predict the lips pose $\hat{\mathbf{z}}_{lips}$ from audio and source lip latents, then pass it to Stage~1.


\begin{table*}[t]
\centering
\resizebox{\textwidth}{!}{
\begin{tabular}{lcccccccccc}
\toprule
\textbf{Model} & \textbf{FID} $\downarrow$ & \textbf{FVD} $\downarrow$ & \textbf{HyperIQA} $\uparrow$ & \textbf{VBench} $\uparrow$  & \textbf{LipScore} $\uparrow$ & \textbf{ID} $\uparrow$ & \textbf{PSNR} $\uparrow$ & \textbf{SSIM} $\uparrow$ & \textbf{LPIPS} $\downarrow$ \\
\midrule
\rowcolor{black!5}
\multicolumn{10}{c}{\normalsize \textbf{Reconstruction}} \\
DiffDub~\cite{liu2024diffdub} & 16.11 & 120.24 & 57.91 & 0.673 & 0.37 & 0.58 & 26.65 & 0.89 & 0.114 \\
Diff2Lip~\cite{mukhopadhyay2024} & 8.76 & 57.96 & 68.31 & 0.672 & 0.47 & 0.79 & 25.64 & 0.94 & 0.056 \\
TalkLip~\cite{wang2023seeing} & 13.97 & 78.66 & 64.52 & 0.667 & 0.57 & 0.79 & 32.03 & 0.94 & 0.062 \\
LatentSync~\cite{li2025latentsync} & 5.30 & 36.47 & 73.10 & \underline{0.682} & 0.55 & \textbf{0.86} & \textbf{33.61} & \textbf{0.97} & \textbf{0.015} \\
IP-LAP~\cite{Zhong_2023_CVPR} & 7.91 & 39.89 & 69.57 & 0.674 & 0.40 & 0.82 & \underline{32.97} & \underline{0.95} & 0.033 \\
KeySync~\cite{bigata2025keysync} & 5.48 & 24.80 & \textbf{74.18} & 0.681 & 0.56 & 0.81 & 30.39 & 0.93 & 0.030 \\
\rowcolor{blue!10}
FlashLips -- U-Net (Ours) & \underline{4.75} & \underline{15.20} & 73.81 & \textbf{0.687} & \underline{0.70} & \underline{0.85} & 32.86 & 0.94 & 0.022 \\
\rowcolor{blue!10}
FlashLips -- Transformer (Ours) & \textbf{4.43} & \textbf{12.31} & \underline{74.06} & \textbf{0.687} & \textbf{0.71} & \textbf{0.86} & 32.88 & 0.94 & \underline{0.021} \\
\midrule
\rowcolor{black!5}
\multicolumn{10}{c}{\normalsize \textbf{Cross-Audio}} \\
DiffDub~\cite{liu2024diffdub} & 18.31 & 123.34 & 57.91 & 0.668 & 0.30 & 0.59 & -- & -- & -- \\
Diff2Lip~\cite{mukhopadhyay2024} & 9.55 & 80.43 & 69.01 & 0.667 & 0.31 & 0.76 & -- & -- & -- \\
TalkLip~\cite{wang2023seeing} & 17.64 & 134.70 & 64.62 & 0.645 & 0.21 & 0.74 & -- & -- & -- \\
LatentSync~\cite{li2025latentsync} & 7.69 & 46.08 & 72.52 & 0.680 & 0.33 & \textbf{0.84} & -- & -- & -- \\
IP-LAP~\cite{Zhong_2023_CVPR} & 9.05 & 43.13 & 69.49 & 0.670 & 0.25 & \underline{0.81} & -- & -- & -- \\
KeySync~\cite{bigata2025keysync} & 6.81 & 37.55 & \textbf{74.13} & 0.676 & 0.36 & 0.79 & -- & -- & -- \\
\rowcolor{blue!10}
FlashLips -- U-Net (Ours) & \underline{6.23} & \underline{33.57} & 73.58 & \underline{0.681} & \textbf{0.38} & \underline{0.81} & -- & -- & -- \\
\rowcolor{blue!10}
FlashLips -- Transformer (Ours) & \textbf{5.89} & \textbf{29.40} & \underline{73.84} & \textbf{0.682} & \underline{0.37} & \underline{0.81} & -- & -- & -- \\
\bottomrule
\end{tabular}
}
\caption{\textbf{Quantitative Comparison.} Comparison on reconstruction and cross-audio scenarios over 100 randomly sampled reconstruction videos and 100 cross-audio pairs from HDTF, CelebV-HQ, and CelebV-Text. Best results are \textbf{bold}; second-best are \underline{underlined}.}
\label{tab:main_metrics}
\end{table*}

\begin{table}[t]
\centering
\begin{tabular}{lcc}
\toprule
\textbf{Model} & \textbf{FPS} $\uparrow$ & \textbf{Speedup $\times$} \\
\midrule
DiffDub~\cite{liu2024diffdub} & 1.86 & 58.8 \\
Diff2Lip~\cite{mukhopadhyay2024} & 19.77 & 5.5 \\
TalkLip~\cite{wang2023seeing} & 51.53 & 2.1 \\
LatentSync~\cite{li2025latentsync} & 5.70 & 19.2 \\
IP-LAP~\cite{Zhong_2023_CVPR} & 4.24 & 25.8 \\
KeySync~\cite{bigata2025keysync} & 3.60 & 30.4 \\
\rowcolor{blue!10}
FlashLips -- U-Net (Ours) & \textbf{109.41} & 1.0 \\
\rowcolor{blue!10}
FlashLips -- Transformer (Ours) & \underline{66.84} & 1.6  \\
\bottomrule
\end{tabular}
\caption{\textbf{Inference Speed.} Speed comparison in frames per second (FPS). “Speedup” denotes the inference speed gain of our fastest model (FlashLips -- U-Net) over each method. Measured on the same clip: 5 warm-ups, then 10 runs to average FPS.
}
\label{tab:speed}
\end{table}

\begin{table}[t]
\centering
\footnotesize
\setlength{\tabcolsep}{3pt}       
\renewcommand{\arraystretch}{0.85}
\resizebox{0.95\linewidth}{!}{
\begin{tabular}{@{}lccccc@{}}
\toprule
\textbf{\# Ref. Lats} & \textbf{FVD} $\downarrow$ & \textbf{HyperIQA} $\uparrow$ & \textbf{LipScore} $\uparrow$ & \textbf{ID} $\uparrow$ & \textbf{PSNR} $\uparrow$ \\
\midrule
\rowcolor{black!5}\multicolumn{6}{c}{\footnotesize \textbf{Reconstruction}} \\
1  & 12.53 & 74.05 & 0.69 & \underline{0.85} & 32.71 \\
4  & 12.31 & 74.06 & 0.71 & \textbf{0.86}    & 32.88 \\
8  & 12.47 & \underline{74.07} & 0.73 & \underline{0.85} & 33.00 \\
16 & \textbf{11.90} & \textbf{74.08} & \underline{0.74} & \textbf{0.86} & \underline{33.02} \\
32 & \underline{12.16} & \textbf{74.08} & \textbf{0.75} & \textbf{0.86} & \textbf{33.10} \\
\midrule
\rowcolor{black!5}\multicolumn{6}{c}{\footnotesize \textbf{Cross-Audio}} \\
1  & 41.38 & 73.80 & \textbf{0.40} & 0.79 & -- \\
4  & 29.40 & 73.84 & \underline{0.37} & \underline{0.81} & -- \\
8  & 29.54 & \underline{73.88} & 0.35 & \underline{0.81} & -- \\
16 & \underline{26.35} & \underline{73.88} & 0.34 & \underline{0.81} & -- \\
32 & \textbf{25.16} & \textbf{73.89} & 0.32 & \textbf{0.82} & -- \\
\bottomrule
\end{tabular}%
}
\caption{\textbf{Reference Latent Ablation (Transformer).} Ablation of the number of reference latents for the Transformer base model on a subset of metrics. Full ablations are in \Cref{tab:combined_refs_models_ablation}.}
\label{tab:ablation_transformer_updated}
\end{table}

\begin{table}[t]
\centering
\resizebox{\linewidth}{!}{
\begin{tabular}{lccccc}
\toprule
\textbf{\# Ref. Lats} & \textbf{FVD} $\downarrow$ & \textbf{HyperIQA} $\uparrow$ & \textbf{LipScore} $\uparrow$ & \textbf{ID} $\uparrow$ & \textbf{PSNR} $\uparrow$ \\
\midrule
\rowcolor{black!5}
\multicolumn{6}{c}{\normalsize \textbf{Reconstruction}} \\
1  & 15.68 & 73.79 & 0.67 & \textbf{0.85} & 32.74 \\
4  & \underline{15.20} & 73.81 & \underline{0.70} & \textbf{0.85} & 32.86 \\
8  & 15.61 & \underline{73.82} & \underline{0.70} & \textbf{0.85} & 32.92 \\
16 & \textbf{15.07} & \textbf{73.83} & \textbf{0.71} & \textbf{0.85} & \underline{32.95} \\
32 & 15.85 & \textbf{73.83} & 0.69 & \textbf{0.85} & \textbf{32.97} \\
\midrule
\rowcolor{black!5}
\multicolumn{6}{c}{\normalsize \textbf{Cross-Audio}} \\
1  & 42.78 & 73.51 & \textbf{0.40} & 0.79 & -- \\
4  & 33.57 & 73.58 & \underline{0.38} & \textbf{0.81} & -- \\
8  & 32.64 & \underline{73.61} & 0.36 & \underline{0.80} & -- \\
16 & \underline{31.34} & \textbf{73.63} & 0.34 & \textbf{0.81} & -- \\
32 & \textbf{28.54} & \textbf{73.63} & 0.32 & \textbf{0.81} & -- \\
\bottomrule
\end{tabular}
}
\caption{\textbf{Reference Latent Ablation (U-Net).} Ablation of the number of reference latents for the U-Net base model on a subset of metrics. Full ablations are provided in \Cref{tab:combined_refs_models_ablation}.}
\label{tab:ablation_unet_updated}
\end{table}

\begin{table}[t]
\centering
\footnotesize
\setlength{\tabcolsep}{4pt}
\resizebox{\columnwidth}{!}{%
\begin{tabular}{lcccc}
\toprule

\multirow{2}{*}{\textbf{Lips Encoder Variant}}
  & \textbf{PSNR} $\uparrow$ 
  & \textbf{SSIM} $\uparrow$
  & \textbf{LipScore} $\uparrow$ 
  & \textbf{ID} $\uparrow$ \\

\cmidrule(lr){2-3}
\cmidrule(lr){4-5}

& \multicolumn{2}{c}{\textbf{Reconstruction}}
& \multicolumn{2}{c}{\textbf{Cross-Audio}} \\

\midrule

V1 2D   & 27.11 & 0.88 & 0.30 & \textbf{0.83} \\
V1 4D   & 30.74 & 0.91 & 0.32 & \textbf{0.83} \\
V1 8D   & 31.63 & 0.93 & 0.34 & \textbf{0.83} \\
V1 16D  & 31.95 & 0.93 & 0.34 & \underline{0.82} \\

\midrule

V1 8D + V2 2D   & 32.16 & 0.93 & 0.36 & 0.81 \\
\rowcolor{blue!10}
V1 8D + V2 4D   & 32.86 & 0.94 & \underline{0.38} & 0.80 \\
V1 8D + V2 8D   & \underline{34.86} & \underline{0.95} & \underline{0.38} & 0.74 \\
V1 8D + V2 16D  & \textbf{36.86} & \textbf{0.97} & \textbf{0.40} & 0.63 \\

\bottomrule
\end{tabular}
}
\caption{\textbf{Lips Encoder Ablation.} V1 (frozen expression encoder) saturates near 8D; adding a lips-crop residual (V2) improves reconstruction but reduces cross-audio ID. The 12D setting (V1 8D + V2 4D) offers the best quality–disentanglement trade-off.}
\label{tab:lips_encoder_ablation}
\end{table}

\begin{figure}[t]
  \centering
  \includegraphics[width=\columnwidth]{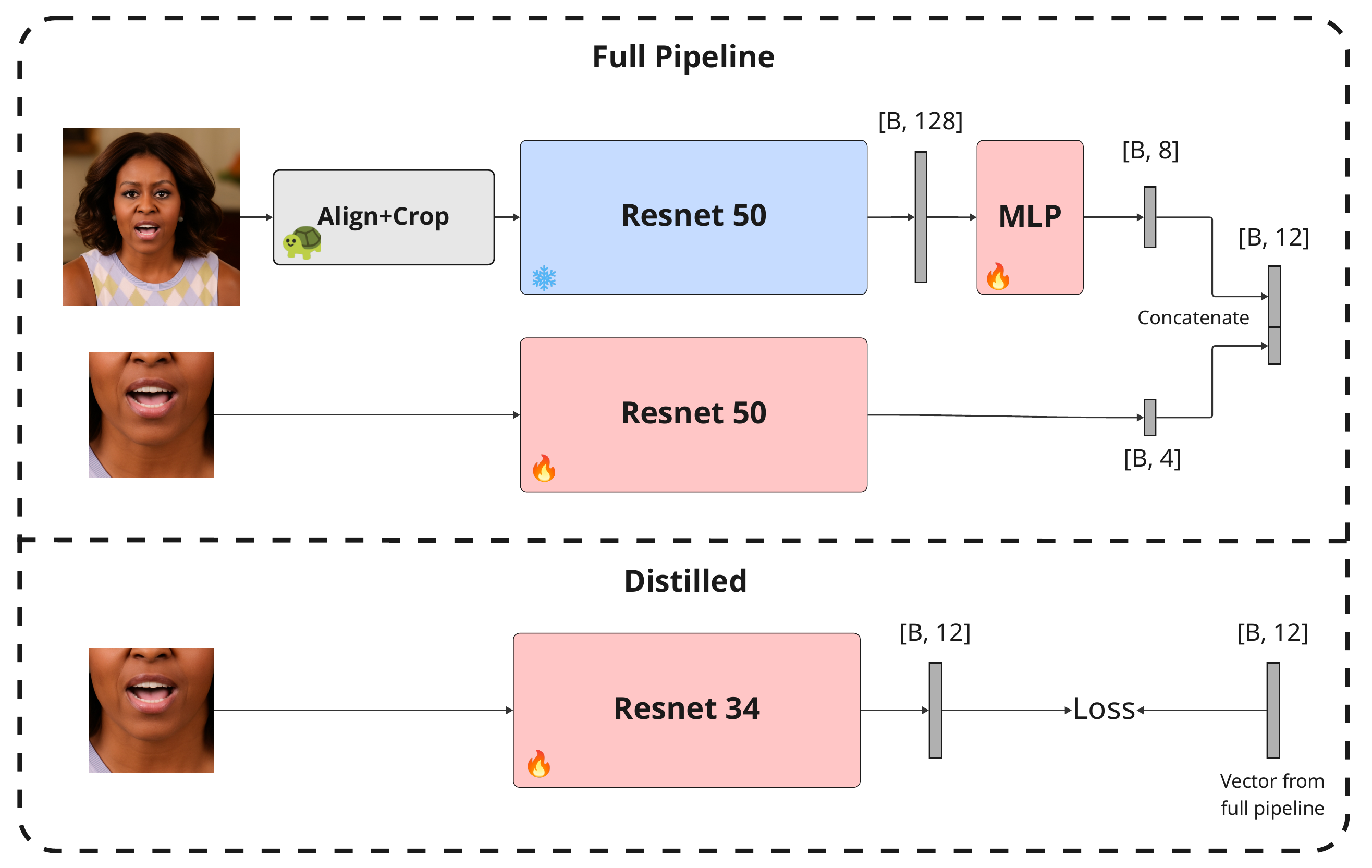}
  \caption{\textbf{Lips Encoder.} A frozen expression encoder with an MLP projector and a mouth-crop CNN produce an 8D+4D lips vector. A distilled ResNet-34 replicates this mapping on inference.}
  \label{fig:lips_encoder}
\end{figure}

\section{Experiments}
\label{sec:experiments}
We evaluate \emph{FlashLips} end-to-end under two standard protocols: (i) \emph{reconstruction} (same-clip audio) and (ii) \emph{cross-audio} (driven by a different clip). For both settings, we report lip–audio synchronization, visual quality, identity preservation, and runtime efficiency. The full set of metrics is defined in \cref{ssec:eval_metrics}. Unless noted otherwise, implementation details (model sizes, preprocessing, schedules) are in the supplementary.

\subsection{Datasets}
A key advantage of our design is that training the most compute-heavy component (Stage~1) does \emph{not} require audio or even lip-sync supervision: we only need images with a visible mouth. This allows us to leverage substantially broader data than typical lip-sync pipelines.

\noindent\textbf{Stage~1 (image-only).}
We train the reconstruction/self-refinement editor on MEAD~\cite{kaisiyuan2020mead}, FEED~\cite{drobyshev2024emoportraits}, HDTF~\cite{zhang2021flow}, NeRSemble~\cite{kirschstein2023nersemble}, CelebV-Text~\cite{yu2022celebvtext}, CelebV-HQ~\cite{zhu2022celebvhq}, and a private collection of controlled laboratory recordings with diverse facial expressions (\(\sim\)500 hours).

\noindent\textbf{Stage~2 (audio-visual).}
For audio-to-lips, we use Hallo3~\cite{cui2024hallo3}, RAVDESS~\cite{livingstone2018ryerson}, MEAD, HDTF, NeRSemble, CelebV-Text, CelebV-HQ, and our private dataset. Stage~2 is trained to predict the lips vector from speech features.

For quantitative evaluation, we randomly sample 100 reconstruction videos and 100 cross-audio pairs from HDTF, CelebV-HQ and CelebV-Text eval sets.

\subsection{Training Details}

\paragraph{Stage~1 (Reconstruction \& LipsChange).}
For the one-step editor, we have two backbone versions: a U-Net with 250M parameters and a ViT-like visual transformer with 300M. Both are optimized with AdamW and a OneCycleLR schedule: the learning rate starts at \(1\!\times\!10^{-4}\) and anneals to \(1\!\times\!10^{-8}\). We train the reconstruction phase for \(1\) million iterations on \(8\times\) NVIDIA H100 80GB HBM3 GPUs with a total batch size of \(32\). After convergence, we train the mask-free self-refinement \textit{LipsChange} (\cref{sssec:maskfree}) using symmetric \emph{source~\(\leftrightarrow\)~changed} pseudo-pairs for 200k iterations.

\vspace{-10pt}

\paragraph{Lips Encoder and Distillation.}
As shown in \cref{fig:lips_encoder}, our \emph{Lips Encoder} starts from a frozen expression encoder \cite{drobyshev2024emoportraits} that maps a face crop to an expression latent, followed by a small trainable head (ResNet-50 + MLP) that projects to the compact lips-pose vector used by Stage~1. We train this head during the Stage~1 reconstruction phase. To remove pre-processing overhead at inference (the expression backbone requires several alignment/cropping steps), we \emph{distill} the pipeline into a single ResNet-50 that consumes the full image and predicts the same lips-pose vector. Distillation is performed by matching the outputs of the full pipeline over the last \(200\text{k}\) iterations of Stage~1 reconstruction training.  

\vspace{-10pt}

\paragraph{Stage~2 (Audio-to-Lips).}
Stage~2 is a transformer conditioned on wav2vec 2.0 features, emotion class inferred from audio, and $K$ random lip latents and trained with a conditional flow-matching objective in the lips-pose vector space (\cref{sec:stage2}). We use the same lips-pose vector definition as Stage~1 to ensure compatibility. For training, we utilize the AdamW optimizer with default parameters and a constant $5 \times 10^{-5}$ learning rate. The transformer backbone has 150M parameters. Additional architectural details are provided in the supplementary.

\subsection{Evaluation Metrics}
\label{ssec:eval_metrics}
We evaluate models using a diverse set of metrics that target lip–audio synchronization, visual quality, identity preservation, and efficiency.

To assess lip–audio synchronization accuracy, we use LipScore~\cite{bigata2025keyface}. For reconstruction fidelity and perceptual quality, we adopt both reference-based and no-reference metrics: Fréchet Inception Distance (FID)~\cite{heusel2017} at the frame level and Fréchet Video Distance (FVD)~\cite{unterthiner2019} at the video level for overall visual quality and temporal consistency; LPIPS~\cite{zhang2018perceptual} as a reference-based perceptual similarity measure; and HyperIQA~\cite{Su_2020_CVPR} as a no-reference image-quality metric correlated with human perception. We also report Peak Signal-to-Noise Ratio (PSNR) and Structural Similarity Index (SSIM)~\cite{wang2004} for pixel-level structural similarity.

To evaluate identity preservation, we define an identity (ID) metric as the cosine similarity between predicted and ground-truth face embeddings computed per frame using FaceNet-512~\cite{schroff2015, serengil2024lightface}. For overall visual consistency, we report VBench~\cite{huang2023vbench} scores: Subject Consistency (SC), Background Consistency (BC), Motion Smoothness (MS), Dynamic Degree (DD), Aesthetic Quality (AQ), Imaging Quality (IQ), and an aggregated “Total” - the mean of the normalized metrics.

Finally, to measure efficiency, we report inference speed in frames per second (FPS). For each model, the same clip is processed 5 times for warm-up and 10 times for measurement, and we report the average runtime over the measured runs.

\subsection{Quantitative Evaluation}
\label{ssec:quant_eval}
We evaluate reconstruction and cross-audio performance against recent state-of-the-art methods, including DiffDub~\cite{liu2024diffdub}, Diff2Lip~\cite{mukhopadhyay2024}, TalkLip~\cite{wang2023seeing}, LatentSync~\cite{li2025latentsync}, IP-LAP~\cite{Zhong_2023_CVPR}, and KeySync~\cite{bigata2025keysync}. Results are shown in ~\Cref{tab:main_metrics,tab:speed}, with full VBench metrics in \Cref{tab:vbench_metrics}.
 
Both our FlashLips models achieve the best overall FID and FVD in both protocols, while also being substantially faster than the next-best method (up to \textbf{30.4$\times$} vs.\ KeySync). On HyperIQA, we rank second and third with only marginal differences from the top model, a metric commonly used as a proxy for human perceptual quality, and obtain the highest VBench total score, indicating strong perceptual quality, temporal consistency, and background preservation.

In LipScore, our models are the top two in both protocols, ensuring highest lip-sync quality. In identity preservation (ID), we are first in reconstruction tied with LatentSync, and second in cross-audio behind it, tied with IP-LAP, while being up to \textbf{19.2$\times$} and \textbf{25.8$\times$} faster, respectively. This reflects strong lip–audio alignment and identity retention across the full sequence.

For pixel-level metrics (PSNR, SSIM), we place third with a narrow gap, and second  in LPIPS, showing that our one-step reconstruction maintains high image fidelity.

Overall, FlashLips delivers state-of-the-art visual and synchronization quality while maintaining faster-than-real-time throughput: \textbf{109.4}~FPS for the U-Net and \textbf{66.8}~FPS for the Transformer base models. \textit{User study results are provided in the supplementary materials.}

\subsection{Qualitative Evaluation}
\label{ssec:qual_eval}
Qualitative results in \cref{fig:qualitative_cross} and \cref{fig:qualitative_recon} show that both FlashLips variants produce visually realistic, temporally coherent animations with strong identity preservation. Compared to existing approaches, our method yields smoother and more natural mouth motion without introducing artifacts or over-exaggerated animation. LatentSync and IP-LAP maintain identity reasonably well but often show weak lip–audio synchronization and noticeable visual artifacts; Diff2Lip exhibits similar issues. DiffDub and Wav2Lip exhibit identity drift and inconsistent lip motion, while KeySync struggles with large head poses, causing deformation and temporal instability. In contrast, FlashLips remains robust across more diverse viewing angles (\cref{fig:qualitative_angles}) and maintains stable lip–audio alignment. It also shows good generalization to out-of-distribution and non-human identities (\cref{fig:qualitative_ood}).

\section{Ablation Study}
\label{sssec:ablaton}
\paragraph{Lips Encoder design and dimensionality.}
Our goal is a lips vector that (a) captures \emph{configuration} of the lips and jaw and is easy to predict from audio, while (b) carrying minimal appearance information (skin tone, lip/teeth geometry) that should instead come from the reference identity latent. Lower appearance leakage simplifies Stage~2 (which cannot infer appearance from audio) and stabilizes training. It also improves pseudo-pair generation for self-refinement: with a well-disentangled vector, we can borrow lips-poses across identities to increase pose diversity.

We experiment with two principal variants. \textbf{(V1)} Use only the frozen expression encoder from \cite{drobyshev2024emoportraits} followed by a small projection head. \textbf{(V2)} Augment \emph{V1} with a small \emph{residual} predicted from a lips crop (extracted via face parsing) by a lightweight ResNet. \Cref{tab:lips_encoder_ablation} shows that the \textbf{(V1)} variant already performs strongly with low identity leakage; reconstruction gains saturate around a lips vector of size~8. Adding a residual progressively improves reconstruction, but hurts disentanglement, especially after the residual reaches size 8. To test leakage, we substituted lips-pose vectors from different identities. We therefore adopt a \textbf{12D} mouth vector with an \(8\)D encoder component + \(4\)D residual, which offers a good trade-off: high reconstruction quality and low identity leakage.

\twocolumn[{%
\begin{center}
    \captionsetup{type=figure}
    \includegraphics[width=0.75\textwidth]{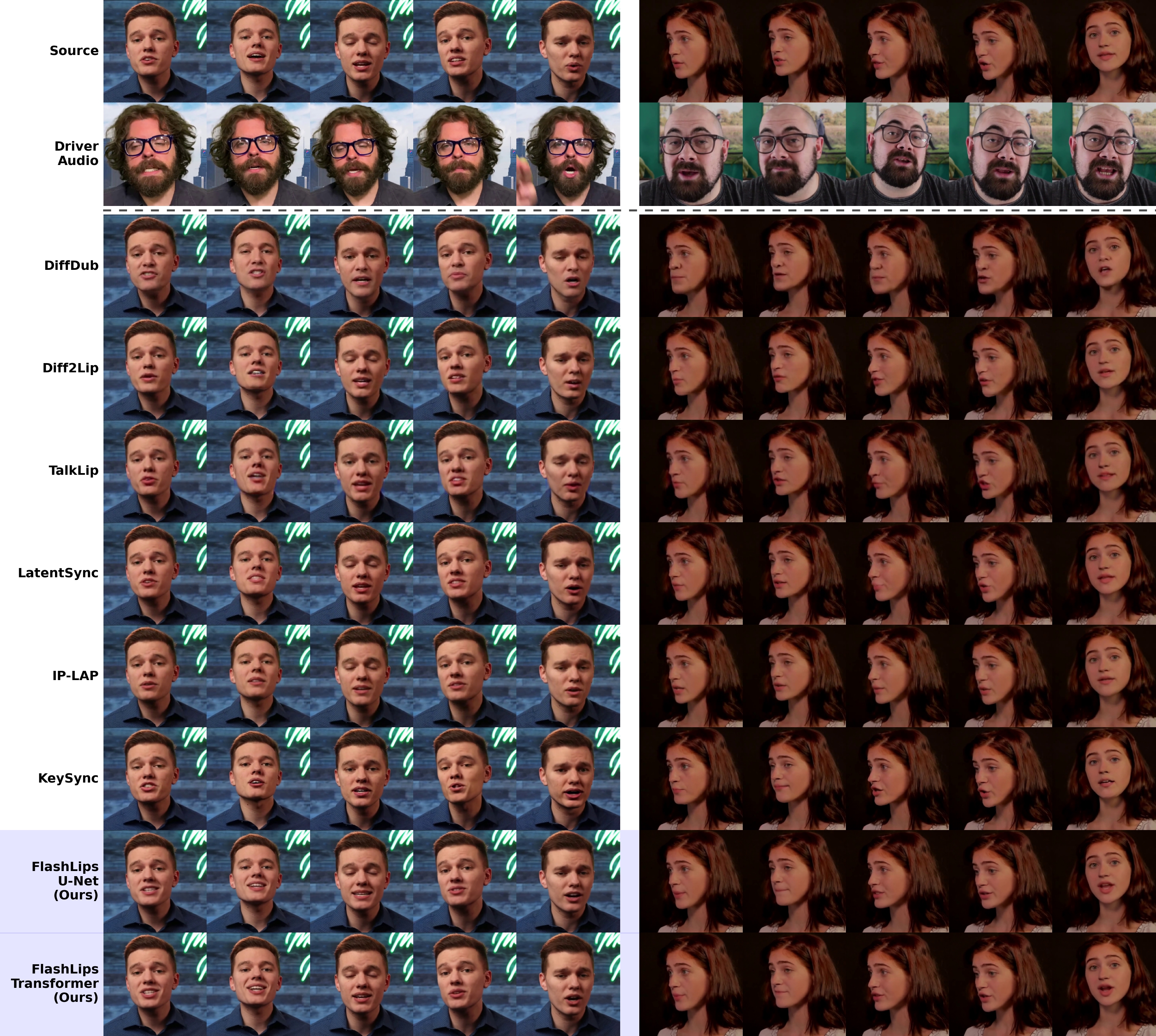}
    \caption{\textbf{Qualitative Comparison -- Cross Audio}. Comparison with other lip-sync methods for cross-audio. The top two rows show the source and audio-driving videos, followed by lip-synced outputs from each method.}
    \label{fig:qualitative_cross}
\end{center}%
}]

\paragraph{Number of reference frames.}
We vary the number of reference lips-pose vectors used in Stage~2. As shown in \Cref{tab:ablation_transformer_updated,tab:ablation_unet_updated}, moving from \(1\) to \(4\) references improves identity preservation with negligible impact on sync; beyond \(4\) gains are marginal and we occasionally observe sync degradation—likely because too much identity information reduces the work done by audio-to-lips network and increases sensitivity when references come from a different clip at inference. We therefore use \textbf{4 reference frames}.

\paragraph{Stage~1 backbone: U-Net vs. Transformer.}
Both backbones have similar accuracy (\Cref{tab:main_metrics}): the transformer is slightly better perceptually but slower, while the U-Net is faster. We use the U-Net for speed and the transformer for higher visual quality; architectural details are in the supplementary.

\paragraph{Summary.}
The ablations support three design choices used in the main results: a 12D lips-pose vector (8D encoder + 4D residual), four reference frames, and either a U-Net (speed) or transformer (quality) backbone for Stage~1.

\section{Conclusion}
\label{sec:conclusion}
\vspace{-0.35em}
FlashLips shows that lip-sync can be posed as \emph{deterministic editing}: a one-step, reconstruction-only latent editor driven by a flow-matched audio-to-lips module, with self-refinement removing the need of masks at inference. This yields a simple, modular system that runs faster than real time ($>$100 FPS with the U-Net variant) while matching or surpassing recent diffusion/GAN baselines in visual quality, identity retention, and lip–audio alignment. Our analysis indicates that a compact, disentangled control vector plus self-refinement are sufficient to localize edits to the mouth, avoid identity/background drift, and decouple \emph{what} to render from \emph{how} to render, enabling practical dubbing and production use. Future work includes improving robustness under occlusions and extreme motion, and enriching the control space with prosody- and emotion-aware signals.

\section{Acknowledgments}
\label{sec:acknowledgments}
\vspace{-0.35em}
This work was conducted at Cantina Labs using their resources. We thank Ruslan Vorovchenko, Kacper Kania, Varun Menon, and Jenya Chelishev for their assistance.

\clearpage
{
    \small
    \bibliographystyle{ieeenat_fullname}
    \bibliography{main}
}

\clearpage
\setcounter{page}{1}
\maketitlesupplementary

\setcounter{section}{0}
\renewcommand\thesection{\Alph{section}}
\renewcommand\thesubsection{\thesection.\arabic{subsection}}
\renewcommand\thefigure{\thesection.\arabic{figure}}
\renewcommand\thetable{\thesection.\arabic{table}}
\renewcommand\theequation{\thesection.\arabic{equation}}
\setcounter{figure}{0}
\setcounter{table}{0}
\setcounter{equation}{0}

\section{Training Details}
\subsection{Data Augmentation}
During training for both stages, we apply the following augmentations. All images are normalized by dividing by 255 to map pixel values into the range \([0, 1]\). For Stage~1, we additionally apply random horizontal flips with probability 0.5 and ColorJitter with a coefficient of 0.4 for hue, brightness, and contrast, 3.2 for saturation, and an overall application probability of 0.9. For the distilled mouth-latent network, we downscale the source image to \(384 \times 384\) using interpolation.

\subsection{Mask Removal Training Details}
\label{app:maskfree_training}
After the reconstruction editor $R_{\phi}$ converges, we synthesize lip-altered counterparts for real frames. 
Given a real frame $S$ and a sampled lips-pose vector $\mathbf{z}_{\text{lips}}$, we produce 
$\tilde{S}=R_{\phi}(S;\mathbf{z}_{\text{lips}})$ and form symmetric pseudo–pairs
$(S\!\rightarrow\!\tilde{S})$ and $(\tilde{S}\!\rightarrow\!S)$. 
We initialize the \emph{LipsChange} editor $L_{\theta}\!\leftarrow\!R_{\phi}$ and fine–tune it on these pairs using the same objective as in \Cref{sssec:losses}. 
At test time, $L_{\theta}$ runs \emph{without} any explicit mouth masks.

\paragraph{Why two directions?}
\begin{itemize}
\item \textbf{$(S\!\rightarrow\!\tilde{S})$ (real $\to$ synth):} input matches inference (real frames), which preserves lip–audio sync; however, the target $\tilde{S}$ may contain minor artifacts from $R_{\phi}$, so training \emph{only} on this direction can reproduce them.
\item \textbf{$(\tilde{S}\!\rightarrow\!S)$ (synth $\to$ real):} target is clean (real $S$), which discourages artifacts; but the input is synthetic and does not match inference, so training \emph{only} on this direction hurts sync/generalization.
\end{itemize}

\paragraph{Mixture.}
We tried various training strategies, but eventually we train our model on a mixture of both directions: sample $(S\!\rightarrow\!\tilde{S})$ with probability $2/3$ and $(\tilde{S}\!\rightarrow\!S)$ with probability $1/3$. 
Empirically, this preserves lip–audio alignment (same LipScore as the real $\to$ synth–only variant) while improving visual quality by avoiding propagation of reconstruction artifacts. 
This self–refinement removes the need for external segmentation at inference and keeps the pipeline mask–free.

\subsection{Architecture Details}
\label{app:arch_details}
All Stage~1 editors (Reconstruction and \emph{LipsChange}) operate on the SDXL VAE latent grid (stride~8). The \emph{input} is the channel-wise concatenation of the masked target latent, the identity-adapted reference latent $f_{\text{ref}}(\mathbf{z}_{\text{ref}})$, and the lips–pose vector tiled to the latent resolution; this totals \textbf{52} channels in our implementation. The network predicts a \textbf{4‑channel latent residual} that is added to the masked latent and decoded by the frozen VAE. \emph{LipsChange} shares the same backbone and is initialized from the Reconstruction network weights for mask‑free self‑refinement.

\vspace{0.5em}
\noindent\textbf{U-Net.}
\Cref{tab:unet_arch} summarizes the U-Net model: we use ResNet2D blocks (GroupNorm (GN) 32, SiLU, $3{\times}3$ convolutions). 
The down path increases the number of channels from \(384 \!\rightarrow\! 512 \!\rightarrow\! 640\) using average-pooling for downsampling; 
the up path mirrors this structure with skip concatenations and resize–conv upsampling, ending with GN+SiLU+Conv to 4 channels. 
This backbone yields the best throughput (see \Cref{tab:speed}) while preserving identity and background consistency and confining edits to the mouth.

\begin{table}[ht]
\centering
\resizebox{\linewidth}{!}{
\begin{tabular}{lcc}
\toprule
\textbf{Stage} & \textbf{Composition} & \textbf{Channels (in $\rightarrow$ out)} \\
\midrule
\textbf{Input} &
Conv2d($k=3, s=1, p=1$) &
52 $\rightarrow$ 384 \\
\midrule
\textbf{Down Block 1} &
4 $\times$ ResNet2D + 1 $\times$ ResNet2D (with AvgPool Downsample) &
384 $\rightarrow$ 384 \\
\textbf{Down Block 2} &
4 $\times$ ResNet2D + 1 $\times$ ResNet2D (with AvgPool Downsample) &
384 $\rightarrow$ 512 \\
\textbf{Down Block 3} &
4 $\times$ ResNet2D (no downsampling) &
512 $\rightarrow$ 640 \\
\midrule
\textbf{Mid Block} &
2 $\times$ ResNet2D &
640 $\rightarrow$ 640 \\
\midrule
\textbf{Up Block 1} &
5 $\times$ ResNet2D + 1 $\times$ ResNet2D (with Upsample) &
1280 $\rightarrow$ 640 \\
\textbf{Up Block 2} &
5 $\times$ ResNet2D + 1 $\times$ ResNet2D (with Upsample) &
1152 $\rightarrow$ 512 \\
\textbf{Up Block 3} &
5 $\times$ ResNet2D (no upsampling) &
896 $\rightarrow$ 384 \\
\midrule
\textbf{Output} &
GN(32) + SiLU + Conv2d($k=3, s=1, p=1$) &
384 $\rightarrow$ 4 \\
\bottomrule
\end{tabular}
}
\caption{
\textbf{Architecture of the U-Net Base Model.} Each ResNet2D block consists of GroupNorm (GN, 32 groups), SiLU activation, and two Conv2d layers ($k=3, s=1, p=1$).
}
\label{tab:unet_arch}
\end{table}

\vspace{0.5em}
\noindent\textbf{Transformer.}
\Cref{tab:transformer_arch} summarizes the ViT-style model: a $1{\times}1$ input projection (52\(\rightarrow\)128), followed by GN and a $1{\times}1$ lift to 1024 channels; then 16 BasicTransformerBlocks (LayerNorm, MHSA with 16 heads \(\times\) 64 dim, GEGLU MLP with \(4\times\) expansion); followed by $1{\times}1$ projections back to 4 output channels. Convolutional pre-/post-projections preserve the 2D grid, while attention improves global consistency at the cost of lower FPS.

\begin{table}[ht]
\centering
\resizebox{\linewidth}{!}{
\begin{tabular}{lcc}
\toprule
\textbf{Stage} & \textbf{Composition} & \textbf{Channels (in $\rightarrow$ out)} \\
\midrule
\textbf{Input Projection} &
Conv2d($k=1, s=1$) &
52 $\rightarrow$ 128 \\
\textbf{Transformer Pre-projection} &
GroupNorm(32) + Conv2d($k=1, s=1$) &
128 $\rightarrow$ 1024 \\
\midrule
\textbf{Transformer Blocks} &
16 $\times$ BasicTransformerBlock: &
1024 $\rightarrow$ 1024 \\
& \quad LayerNorm + MHSA (16 heads, 64-dim/head) & \\
& \quad + LayerNorm + FeedForward (GEGLU, 4$\times$ expansion) & \\
\midrule
\textbf{Transformer Post-projection} &
Conv2d($k=1, s=1$) &
1024 $\rightarrow$ 128 \\
\textbf{Output Projection} &
Conv2d($k=1, s=1$) &
128 $\rightarrow$ 4 \\
\bottomrule
\end{tabular}
}
\caption{
\textbf{Architecture of the Transformer Base Model.} MHSA stands for Multi-Head Self-Attention.
}
\label{tab:transformer_arch}
\end{table}

\vspace{0.5em}
\noindent\textbf{Trade-off.}
Both backbones achieve comparable accuracy (main paper). The transformer is slightly stronger on perceptual metrics, whereas the U-Net is substantially faster. This makes the U-Net preferable for real-time use and the transformer preferable for peak visual quality. The Stage-2 flow-matching transformer (FMT) architecture -- shared by both U-Net and Transformer variants of FlashLips -- is detailed in \Cref{tab:fmt_arch}.

\begin{table}[ht]
\centering
\resizebox{\linewidth}{!}{
\begin{tabular}{lcc}
\toprule
\textbf{Stage} & \textbf{Composition} & \textbf{Dims (in $\rightarrow$ out)} \\
\midrule
\textbf{Input Motion Embedding} &
SequenceEmbed: Linear($d_p \rightarrow d_h$) &
$d_p \rightarrow d_h$ \\
& (no affine norm) & $12 \rightarrow 1024$ \\
\midrule
\textbf{Positional Encoding} &
Fixed sinusoidal encoding (non-learnable), &
$T \times d_h \rightarrow T \times d_h$ \\
& added to token embeddings ($T = 60$ frames) & $60 \times 1024 \rightarrow 60 \times 1024$ \\
\midrule
\textbf{Time Embedding} &
TimestepEmbedder: sinusoidal (256-dim) &
$256 \rightarrow d_h$ \\
& + MLP: Linear($256 \rightarrow d_h$) + SiLU &
$256 \rightarrow 1024$ \\
& + Linear($d_h \rightarrow d_h$) &
$1024 \rightarrow 1024$ \\
\midrule
\textbf{Condition Embedding} &
Concat of identity, audio and emotion latents: &
$(d_{\text{cond}} \rightarrow d_h)$ \\
& $[w_r, w_a, w_e]$ with &
$d_{\text{cond}} = d_p \cdot n_{\text{id}} + d_a + d_e$ \\
& Linear($d_{\text{cond}} \rightarrow d_h$) &
$ (12 n_{\text{id}} + 512 + 7) \rightarrow 1024$ \\
\midrule
\textbf{FMT Blocks} &
$8 \times$ FMTBlock: &
$d_h \rightarrow d_h$ \\
& \quad AdaLN-modulated MHSA &
$1024 \rightarrow 1024$ \\
& \quad (8 heads, 128-dim/head) &
\\
& \quad + AdaLN-modulated MLP &
\\
& \quad MLP: Linear($d_h \rightarrow 4d_h$) + GELU &
$1024 \rightarrow 4096$ \\
& \quad \phantom{MLP:} + Linear($4d_h \rightarrow d_h$) &
$4096 \rightarrow 1024$ \\
& \quad AdaLN MLP: SiLU + Linear($d_h \rightarrow 6d_h$) &
$1024 \rightarrow 6144$ \\
\midrule
\textbf{Output Decoder} &
AdaLN: LayerNorm (no affine) + &
$d_h \rightarrow d_h$ \\
& SiLU + Linear($d_h \rightarrow 2d_h$) &
$1024 \rightarrow 2048$ \\
& Linear($d_h \rightarrow d_w$) &
$1024 \rightarrow 12$ \\
\bottomrule
\end{tabular}
}
\caption{
\textbf{Architecture of the Flow Matching Transformer (FMT).} $d_p$ is the lips-pose latent dimension,
$d_h$ is the hidden size. MHSA stands for Multi-Head Self-Attention.
}
\label{tab:fmt_arch}
\end{table}

\section{User Study}
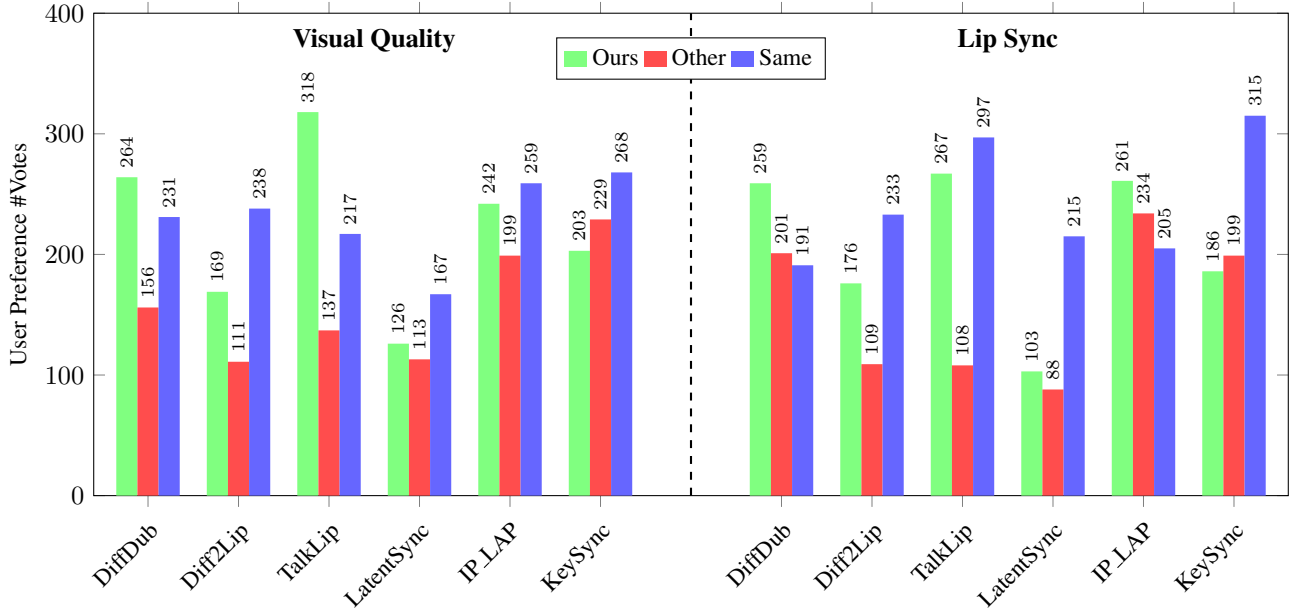
\begin{figure*}[t]
\centering
\begin{tikzpicture}

\begin{axis}[
    width=\textwidth,
    height=8cm,
    ybar=0pt,           
    bar width=8pt,
    ymin=0, 
    ymax=400,           
    enlarge x limits=0.05,
    symbolic x coords={
        DiffDub,Diff2Lip,TalkLip,LatentSync,IP-LAP,KeySync,
        SEP,
        DiffDub2,Diff2Lip2,TalkLip2,LatentSync2,IP-LAP-2,KeySync2
    },
    xtick={
        DiffDub,Diff2Lip,TalkLip,LatentSync,IP-LAP,KeySync,
        DiffDub2,Diff2Lip2,TalkLip2,LatentSync2,IP-LAP-2,KeySync2
    },
    xticklabels={
        DiffDub,Diff2Lip,TalkLip,LatentSync,IP-LAP,KeySync,
        DiffDub,Diff2Lip,TalkLip,LatentSync,IP-LAP,KeySync
    },
    xticklabel style={font=\small, rotate=45, anchor=north east},
    ylabel={User Preference \#Votes},
    ylabel style={font=\small},
    nodes near coords,
    nodes near coords style={font=\scriptsize, rotate=90, anchor=west},
    legend style={at={(0.5,0.95)}, anchor=north, legend columns=3, font=\small},
    legend image code/.code={
        \draw[#1, draw=none] (0cm,-0.1cm) rectangle (0.3cm,0.1cm);
    }
]

\addplot[draw=none, fill=green!50] coordinates {
 (DiffDub,264) (Diff2Lip,169) (TalkLip,318) (LatentSync,126)
 (IP-LAP,242) (KeySync,203)
 (SEP,nan)
 (DiffDub2,259) (Diff2Lip2,176) (TalkLip2,267) (LatentSync2,103)
 (IP-LAP-2,261) (KeySync2,186)
};

\addplot[draw=none, fill=red!70] coordinates {
 (DiffDub,156) (Diff2Lip,111) (TalkLip,137) (LatentSync,113)
 (IP-LAP,199) (KeySync,229)
 (SEP,nan)
 (DiffDub2,201) (Diff2Lip2,109) (TalkLip2,108) (LatentSync2,88)
 (IP-LAP-2,234) (KeySync2,199)
};

\addplot[draw=none, fill=blue!60] coordinates {
 (DiffDub,231) (Diff2Lip,238) (TalkLip,217) (LatentSync,167)
 (IP-LAP,259) (KeySync,268) 
 (SEP,nan)
 (DiffDub2,191) (Diff2Lip2,233) (TalkLip2,297) (LatentSync2,215)
 (IP-LAP-2,205) (KeySync2,315)
};

\legend{Ours, Other, Same}

\draw[dashed, thick] (axis cs:SEP,0) -- (axis cs:SEP,400);

\path (axis cs:DiffDub, 395) -- (axis cs:KeySync, 395) 
      node[midway, font=\bfseries, anchor=north] {Visual Quality};

\path (axis cs:DiffDub2, 395) -- (axis cs:KeySync2, 395) 
      node[midway, font=\bfseries, anchor=north] {Lip Sync};

\end{axis}
\end{tikzpicture}
\caption{\textbf{Human Preference Evaluation.} We conducted a user study comparing FlashLips against randomly selected baseline models. Participants indicated their preference across two criteria: Visual Quality and Lip Sync. The chart displays the number of responses favoring our model (Ours), the competing model (Other), or neither (Same).}
\label{fig:user_study}
\end{figure*}

To complement our quantitative evaluation, we conducted a user study comparing FlashLips with several baseline lip-sync models. Using the same 100 cross-audio videos as in our quantitative experiments, we present participants with two videos per trial: one generated by our method and one by a randomly selected baseline. Users evaluate either \textit{Visual Quality} or \textit{Lip Sync}, choosing the preferred video or indicating that both are of the same quality. We collect up to 700 votes per baseline comparison and setting, which are then aggregated. The results are shown in \Cref{fig:user_study}.

Across nearly all baselines, FlashLips is the clear user preference for both visual quality and lip-sync accuracy, with a substantial portion of responses also indicating comparable quality. We outperform DiffDub, Diff2Lip, TalkLip, and IP-LAP by a large margin, with only a minority of votes favoring the competing models. Against LatentSync, most users judge the outputs to be similar, with a slight preference for our method. KeySync -- a considerably slower (by $\times 30.4$ times, see \Cref{tab:speed}) iterative diffusion model -- shows a negligible advantage with $29.0\%$ vs $32.7\%$ of votes for visual quality and $26.6\%$ vs $28.4\%$ for lip-sync, although the vast majority of users still deem the two videos to be of equal quality with $38.3\%$ and $45.0\%$ of votes in the respective settings. We attribute this small disadvantage to artifacts introduced by the SDXL VAE under certain framings and head poses (see \Cref{ssec:limitations}).

Overall, the study highlights that FlashLips delivers competitive or superior perceptual quality while operating orders of magnitude faster than state-of-the-art diffusion-based approaches.

\section{Additional Quantitative Results}
\label{sec:additional_quant_results}
\subsection{Mask Removal: Quantitative Impact}
\label{app:maskfree_vs_mask}
To isolate the effect of removing explicit mouth masks, \Cref{tab:combined_refs_models_ablation} compares the \emph{Transformer with Mask} to our mask-free editors \emph{Transformer Mask-free} and \emph{U-Net Mask-free} under identical evaluation protocols. We treat reconstruction as a sanity check and focus primarily on cross-audio, which reflects the real use case of our model.

\vspace{1em}
\noindent\textbf{Reconstruction.}
Removing the mask improves both fidelity and lip-sync quality for the Transformer variant: LipScore increases from roughly 0.50 to 0.70–0.75, and all fidelity metrics (FID/FVD, LPIPS, PSNR, SSIM, ID) move in the expected direction, suggesting better distribution match, sharper frames and higher identity similarity. This confirms that mask-free editing can localize mouth modifications without sacrificing reconstruction quality.

\vspace{1em}
\noindent\textbf{Cross-audio.}
Mask removal yields the largest improvements in cross-audio. For the Transformer, FID drops from $\sim$10.2 to $\sim$5.7 and FVD from $\sim$74 to $\sim$25-41, indicating cleaner frames and substantially more stable motion. ID improves from $\sim$0.77-0.79 to $\sim$0.81-0.82, and HyperIQA increases slightly. LipScore remains in the same range of 0.35-0.40, showing that lip–audio alignment is preserved. Qualitatively, mask-free models reduce mouth glitches and flicker while providing more stable backgrounds and facial detail. The mask-free U-Net follows the same trend, with slightly worse FID/FVD but higher throughput.

\vspace{1em}
\noindent\textbf{Takeaway.}
Mask-free self-refinement is a key contributor to the final system: it removes the need for segmentation at inference and consistently improves perceptual quality, temporal smoothness, and identity preservation, while maintaining lip–audio alignment comparable to or better than the masked baseline.

\subsection{VBench Results}
\label{app:vbench_results}
\Cref{tab:vbench_metrics} summarizes VBench scores (see \Cref{ssec:eval_metrics}). Across both reconstruction and cross-audio settings, our mask-free models achieve the highest or near-highest total score, demonstrating strong subject and background consistency, motion smoothness, and perceived visual fidelity.

\begin{table}[ht]
\centering
\setlength{\tabcolsep}{3.5pt}
\resizebox{\columnwidth}{!}{%
\begin{tabular}{lccccccH}
\toprule
\rowcolor{black!5}
\multicolumn{8}{c}{\normalsize \textbf{Reconstruction}} \\
\textbf{Model} & \textbf{SC} $\uparrow$ & \textbf{BC} $\uparrow$ & \textbf{MS} $\uparrow$ & \textbf{DD} $\uparrow$ & \textbf{AQ} $\uparrow$ & \textbf{IQ} $\uparrow$ & \textbf{Total} $\uparrow$ \\
\midrule
DiffDub~\cite{liu2024diffdub} & \underline{0.962} & 0.954 & \textbf{0.992} & 0.670 & 0.505 & 0.661 & 0.673 \\
Diff2Lip~\cite{mukhopadhyay2024} & 0.953 & 0.946 & \textbf{0.992} & 0.653 & 0.557 & 0.633 & 0.672 \\
TalkLip~\cite{wang2023seeing} & 0.952 & 0.942 & \textbf{0.992} & \underline{0.727} & 0.528 & 0.596 & 0.667 \\
LatentSync~\cite{li2025latentsync} & \textbf{0.967} & 0.948 & \underline{0.991} & 0.723 & 0.528 & \textbf{0.671} & \underline{0.682} \\
IP-LAP~\cite{Zhong_2023_CVPR} & 0.961 & 0.945 & \textbf{0.992} & 0.720 & 0.519 & 0.640 & 0.674 \\
KeySync~\cite{bigata2025keysync} & 0.953 & 0.948 & \underline{0.991} & \textbf{0.750} & 0.531 & \underline{0.669} & 0.681 \\
\rowcolor{blue!10}
FlashLips -- U-Net (Ours) & 0.957 & \textbf{0.956} & 0.990 & \textbf{0.750} & \underline{0.559} & 0.667 & \textbf{0.687} \\
\rowcolor{blue!10}
FlashLips -- Transformer (Ours) & 0.957 & \underline{0.955} & 0.990 & \textbf{0.750} & \textbf{0.560} & \underline{0.669} & \textbf{0.687} \\
\midrule\midrule
\rowcolor{black!5}
\multicolumn{8}{c}{\normalsize \textbf{Cross-Audio}} \\
\textbf{Model} & \textbf{SC} $\uparrow$ & \textbf{BC} $\uparrow$ & \textbf{MS} $\uparrow$ & \textbf{DD} $\uparrow$ & \textbf{AQ} $\uparrow$ & \textbf{IQ} $\uparrow$ & \textbf{Total} $\uparrow$ \\
\midrule
DiffDub~\cite{liu2024diffdub} & 0.956 & 0.953 & \textbf{0.992} & 0.624 & 0.506 & 0.660 & 0.668 \\
Diff2Lip~\cite{mukhopadhyay2024} & 0.946 & 0.945 & \underline{0.991} & 0.622 & 0.550 & 0.631 & 0.667 \\
TalkLip~\cite{wang2023seeing} & \underline{0.958} & 0.947 & \textbf{0.992} & 0.420 & 0.527 & 0.596 & 0.645 \\
LatentSync~\cite{li2025latentsync} & \textbf{0.963} & 0.953 & \underline{0.991} & \textbf{0.690} & 0.537 & 0.664 & 0.680 \\
IP-LAP~\cite{Zhong_2023_CVPR} & \textbf{0.963} & 0.945 & \textbf{0.992} & 0.670 & 0.519 & 0.639 & 0.670 \\
KeySync~\cite{bigata2025keysync} & 0.951 & 0.949 & \underline{0.991} & \underline{0.680} & 0.529 & \textbf{0.668} & 0.676 \\
\rowcolor{blue!10}
FlashLips -- U-Net (Ours) & 0.955 & \textbf{0.958} & 0.990 & 0.670 & \underline{0.558} & \underline{0.666} & \underline{0.681} \\
\rowcolor{blue!10}
FlashLips -- Transformer (Ours) & 0.955 & \underline{0.957} & 0.990 & \underline{0.680} & \textbf{0.559} & \textbf{0.668} & \textbf{0.682} \\
\bottomrule
\end{tabular}%
}
\caption{\textbf{Quantitative Comparison on VBench.} Video quality evaluation using VBench \cite{huang2023vbench} metrics on 100 randomly sampled reconstruction videos and 100 cross-audio pairs from HDTF, CelebV-HQ and CelebV-Text. Metrics defined in \Cref{ssec:eval_metrics}.}
\label{tab:vbench_metrics}
\end{table}

\begin{table*}[ht]
\centering
\resizebox{\textwidth}{!}{
\begin{tabular}{llcccccccc}
\toprule
\textbf{Model} & \textbf{\# Ref Lats} & \textbf{FID} $\downarrow$ & \textbf{FVD} $\downarrow$ & \textbf{HyperIQA} $\uparrow$ & \textbf{LipScore} $\uparrow$ & \textbf{ID} $\uparrow$ & \textbf{PSNR} $\uparrow$ & \textbf{SSIM} $\uparrow$ & \textbf{LPIPS} $\downarrow$ \\
\midrule
\rowcolor{black!5}
\multicolumn{10}{c}{\normalsize \textbf{Reconstruction}} \\
\midrule
\makecell[l]{Transformer with Mask} & 1  & 8.06 & 57.80 & 73.26 & 0.50 & 0.81 & 27.68 & 0.90 & 0.044 \\
 & 4  & 8.01 & 57.93 & 73.27 & 0.53 & 0.82 & 27.75 & 0.90 & 0.043 \\
 & 8  & 7.95 & 58.15 & 73.28 & 0.55 & 0.82 & 27.77 & 0.90 & 0.043 \\
 & 16 & 7.97 & 57.78 & 73.29 & 0.56 & 0.82 & 27.79 & 0.90 & 0.043 \\
 & 32 & 7.94 & 57.63 & 73.28 & 0.55 & 0.82 & 27.80 & 0.90 & 0.043 \\
\midrule
\makecell[l]{Transformer Mask-free} & 1  & 4.46 & 12.53 & 74.05 & 0.69 & 0.85 & 32.71 & 0.94 & 0.021 \\
 & 4  & 4.43 & 12.31 & 74.06 & 0.71 & 0.86 & 32.88 & 0.94 & 0.021 \\
 & 8  & 4.41 & 12.47 & 74.07 & 0.73 & 0.85 & 33.00 & 0.94 & 0.021 \\
 & 16 & 4.38 & 11.90 & 74.08 & 0.74 & 0.86 & 33.02 & 0.94 & 0.021 \\
 & 32 & 4.36 & 12.16 & 74.08 & 0.75 & 0.86 & 33.10 & 0.94 & 0.020 \\
\midrule
\makecell[l]{U-Net Mask-free} & 1  & 4.73 & 15.68 & 73.79 & 0.67 & 0.85 & 32.74 & 0.94 & 0.022 \\
 & 4  & 4.75 & 15.20 & 73.81 & 0.70 & 0.85 & 32.86 & 0.94 & 0.022 \\
 & 8  & 4.76 & 15.61 & 73.82 & 0.70 & 0.85 & 32.92 & 0.94 & 0.022 \\
 & 16 & 4.66 & 15.07 & 73.83 & 0.71 & 0.85 & 32.95 & 0.94 & 0.022 \\
 & 32 & 4.70 & 15.85 & 73.83 & 0.69 & 0.85 & 32.97 & 0.94 & 0.022 \\
\midrule
\rowcolor{black!5}
\multicolumn{10}{c}{\normalsize \textbf{Cross-Audio}} \\
\midrule
\makecell[l]{Transformer with Mask} & 1  & 10.24 & 73.91 & 73.16 & 0.40 & 0.77 & -- & -- & -- \\
 & 4  & 9.87 & 68.10 & 73.23 & 0.39 & 0.77 & -- & -- & -- \\
 & 8  & 9.74 & 68.77 & 73.26 & 0.38 & 0.78 & -- & -- & -- \\
 & 16 & 9.67 & 66.17 & 73.25 & 0.35 & 0.79 & -- & -- & -- \\
 & 32 & 9.68 & 64.92 & 73.25 & 0.34 & 0.78 & -- & -- & -- \\
\midrule
\makecell[l]{Transformer Mask-free} & 1  & 6.25 & 41.38 & 73.80 & 0.40 & 0.79 & -- & -- & -- \\
 & 4  & 5.89 & 29.40 & 73.84 & 0.37 & 0.81 & -- & -- & -- \\
 & 8  & 5.81 & 29.54 & 73.88 & 0.35 & 0.81 & -- & -- & -- \\
 & 16 & 5.73 & 26.35 & 73.88 & 0.34 & 0.81 & -- & -- & -- \\
 & 32 & 5.68 & 25.16 & 73.89 & 0.32 & 0.82 & -- & -- & -- \\
\midrule
\makecell[l]{U-Net Mask-free} & 1  & 6.54 & 42.78 & 73.51 & 0.40 & 0.79 & -- & -- & -- \\
 & 4  & 6.23 & 33.57 & 73.58 & 0.38 & 0.81 & -- & -- & -- \\
 & 8  & 6.13 & 32.64 & 73.61 & 0.36 & 0.80 & -- & -- & -- \\
 & 16 & 6.07 & 31.34 & 73.63 & 0.34 & 0.81 & -- & -- & -- \\
 & 32 & 6.13 & 28.54 & 73.63 & 0.32 & 0.81 & -- & -- & -- \\
\bottomrule
\end{tabular}
}
\caption{\textbf{Full Ablation Study.} Ablation study of our mask and mask-free models, and different numbers of references for the audio-to-latent model for reconstruction and cross-audio. Metrics computed on 100 randomly sampled reconstruction videos and 100 cross-audio pairs from HDTF, CelebV-HQ and CelebV-Text.}
\label{tab:combined_refs_models_ablation}
\end{table*}

\begin{figure}[!h]
  \centering
  \includegraphics[width=\columnwidth]{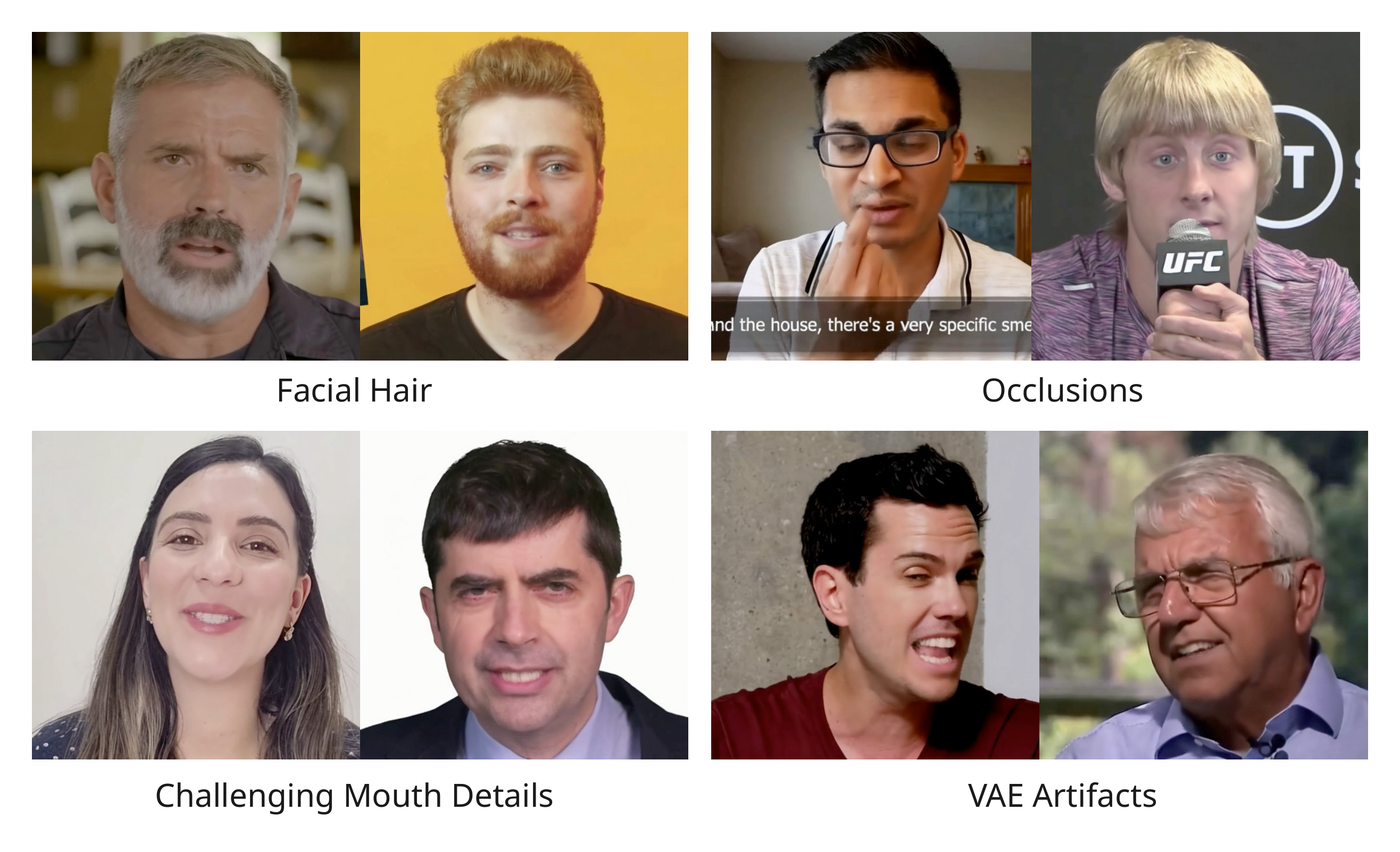}
  \caption{\textbf{Limitations.} Examples illustrating typical failure cases under challenging conditions, including generating facial hair and teeth details, occlusions, and artifacts caused by the SDXL VAE.}
  \label{fig:limitations}
\end{figure}

\section{Limitations}
\label{ssec:limitations}
Although our model produces high-quality lip-sync in most cases, it still exhibits some limitations (\Cref{fig:limitations}). Since the method relies on direct prediction rather than the iterative denoising used in diffusion-based approaches, it can struggle to generate fine-grained facial details, particularly in regions such as facial hair and teeth. While the model was not explicitly trained to handle occlusions, it is often surprisingly robust; however, occluding objects can still degrade lip-sync accuracy in more challenging sequences. A more fundamental limitation stems from the SDXL VAE, whose performance degrades in a predictable manner under certain framings and head poses. The VAE performs well on tight close-ups, but when the subject appears in wider shots, artifacts become more common and can adversely affect the lip-sync quality.

\section{Ethical Considerations and Societal Impact}
Lip-sync technology enables applications such as accessibility tools, film and TV dubbing, translation for non-native audiences, expressive avatars, and content creation. It also carries clear risks: malicious users may create deceptive deepfakes, spread misinformation, or impersonate identities. Our method is intended for beneficial use, and we explicitly discourage any harmful or non-consensual deployment. Any system that alters a person’s likeness should obtain explicit, informed consent.

Our model is trained on publicly available datasets that follow their usage guidelines and on an internal dataset collected with participant consent. As with many audio-visual models, dataset limitations may bring biases across attributes as skin tone, facial structure, language, or accent.

\vspace{0.75em}
\section{Additional Qualitative Results}
\label{sec:additional_qual_results}
\begin{figure*}[!t]
    \centering
    \includegraphics[width=\textwidth]{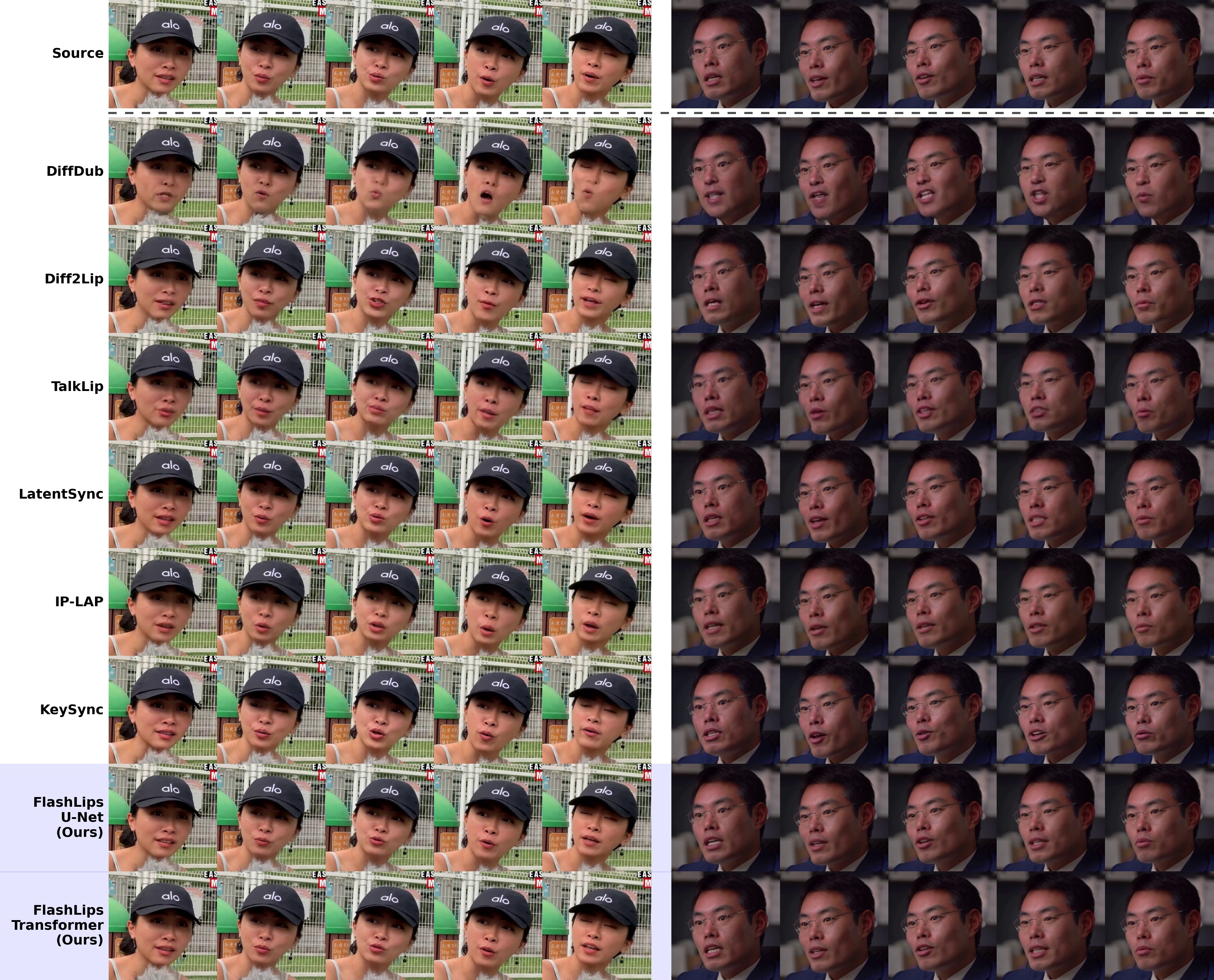}
    \caption{\textbf{Qualitative Comparison -- Reconstruction.} Comparisons with other lip-sync methods for reconstruction. The first row shows the source video; the following rows display the inferred lip-synced videos by each method.}
    \label{fig:qualitative_recon}
\end{figure*}

\begin{figure*}[!t]
    \centering
    \includegraphics[width=0.9\textwidth,
                    trim={10pt 0pt 0pt 0pt},
                    clip]{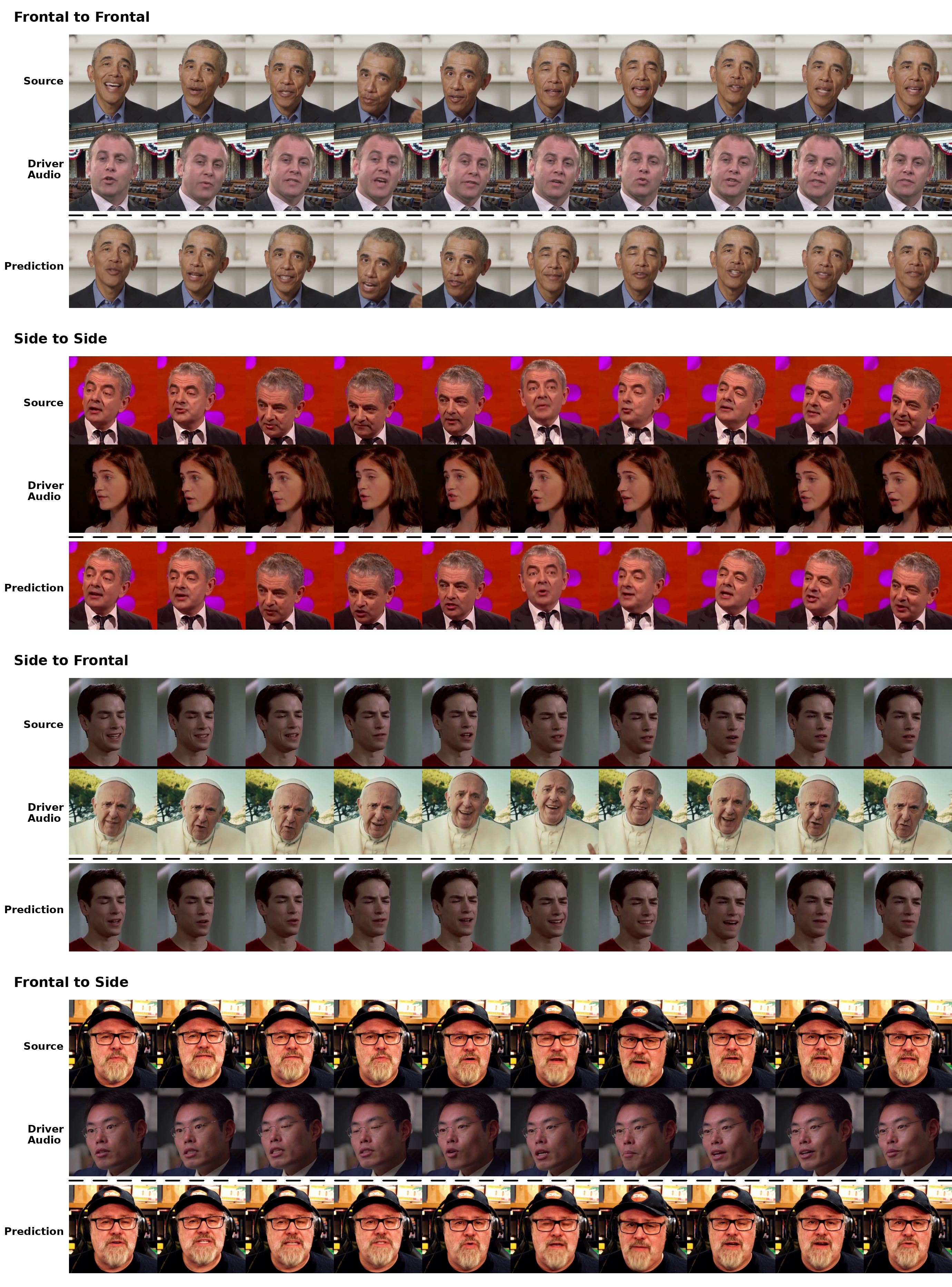}
    \caption{\textbf{Lip-sync results across varying facial pose combinations.} Each triplet shows a source video, video corresponding to the audio driver, and the resulting prediction.}
    \label{fig:qualitative_angles}
\end{figure*}

\begin{figure*}[!t]
    \centering
    \includegraphics[width=\textwidth]{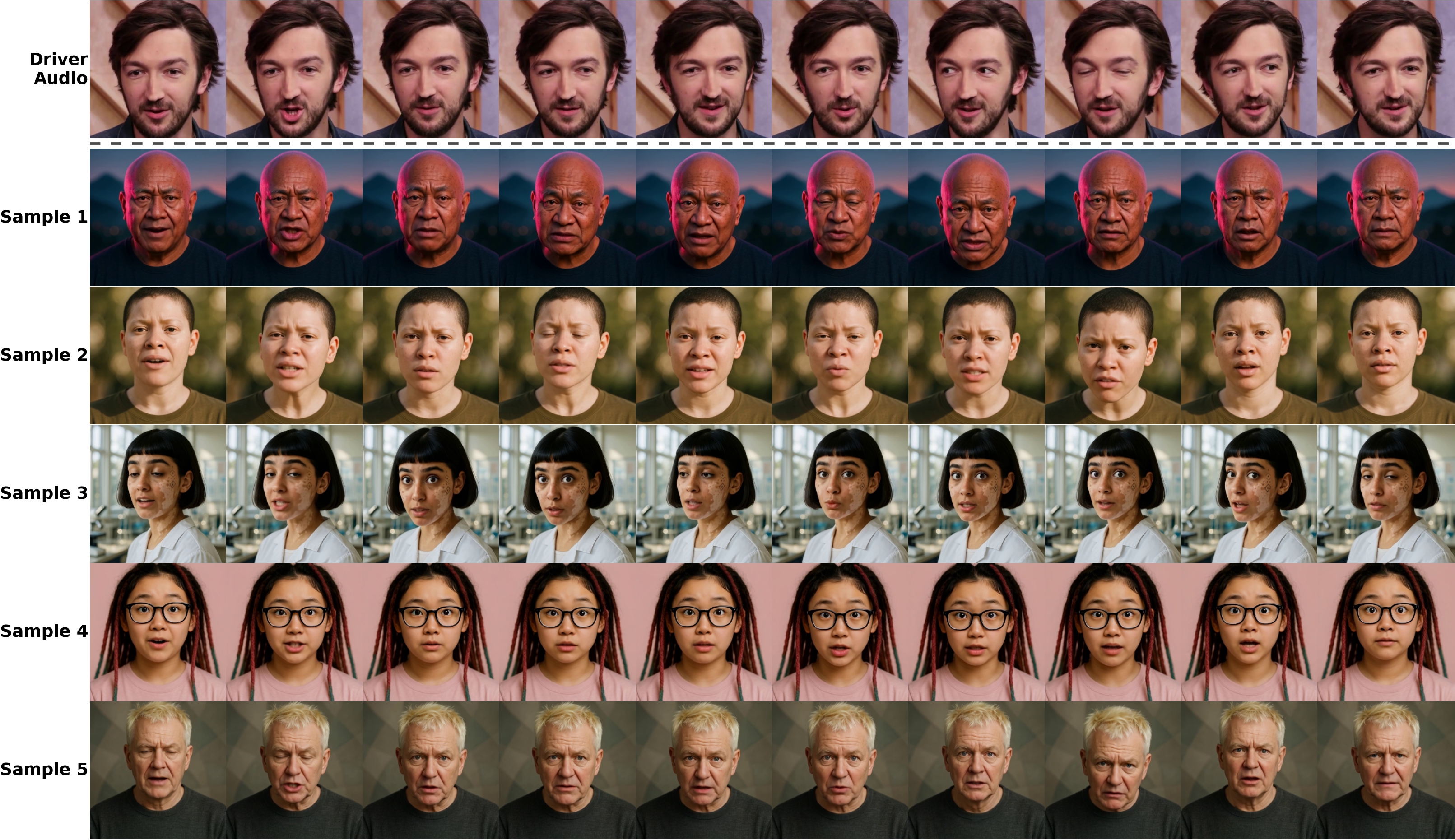}
    \par\vspace{2em}
    \includegraphics[width=\textwidth]{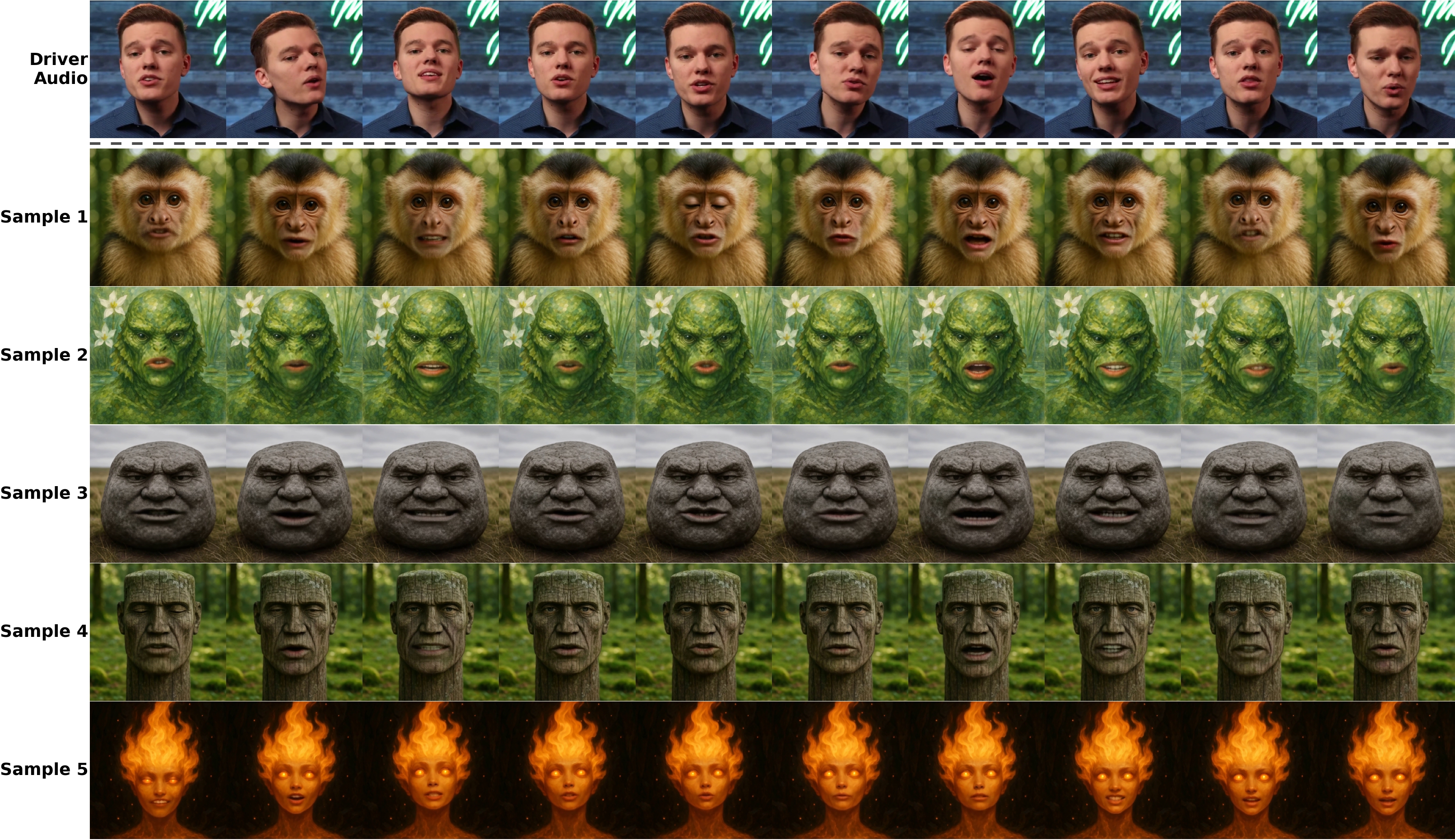}
    \caption{\textbf{Lip-sync results on out-of-distribution (OOD) faces.}
    The top block of images shows results on generated human faces, while the lower block shows results on non-human or stylized faces. 
    Our method maintains consistent lip synchronization and natural articulation across both domains.}
    \label{fig:qualitative_ood}
\end{figure*}

We provide qualitative reconstruction results in \Cref{fig:qualitative_recon}, comparing FlashLips against all baselines. We also assess visual quality across diverse source–driver head-pose combinations, including frontal–frontal, side–side, side–frontal, and frontal–side pairs (\Cref{fig:qualitative_angles}). Finally, we show results on out-of-distribution subjects -- synthetic human faces and non-human or stylized characters -- to demonstrate that our method remains robust and generalizable under these more challenging conditions (\Cref{fig:qualitative_ood}).

\end{document}